%% file: mfgp.tex
\documentclass[12pt]{iopart}
\usepackage{amsgen}
\usepackage{amssymb}
\usepackage{amsbsy}
\usepackage{amsfonts}
\usepackage{pdfpages}
\usepackage{graphicx}
\usepackage{subcaption}
\usepackage{comment}
\usepackage{adjustbox}
\usepackage{booktabs}
\usepackage{array}
\usepackage{float}
\newcommand{\gguide}

\newcommand{\argmax}{\mathop{\mathrm{argmax}}}

\newcommand{\RefEq}[1]{equation~\textup{(\ref{#1})}}

\begin{document}

\title[Multi-fidelity GP surrogate modeling for regression problems in physics]{Multi-fidelity Gaussian process surrogate modeling for regression problems in physics}

\input{sections/authorlist.tex}

\vspace{10pt}
\begin{indented}
\item[] March 2024
\end{indented}

\begin{abstract}
\input{sections/abstract}
\end{abstract}

\textbf{Keywords:} multi-fidelity, machine learning, Gaussian processes, physical simulations
\maketitle

\section{Introduction}
\input{sections/intro}

\section{Background}
\label{sec:theory}
\subsection{Gaussian process regression}
\input{sections/theory/gp}
\subsection{Kernels}
\input{sections/theory/kernels}
\subsection{Calibration}
\label{sec:calibration}
\input{sections/theory/calibration.tex}

\section{Methodology}
\label{sec:method}
\subsection{Multi-fidelity}
\input{sections/theory/mf}

\subsubsection{Linear Auto-regressive model}
\input{sections/theory/ar1}

\subsubsection{Non-linear auto-regressive models}
\label{sec:nonliner-ar}
\input{sections/theory/data_fusion}

\subsubsection{Deep Gaussian Processes}
\label{sec:dgp}
\input{sections/theory/deepGP}

\subsection{Adaptive model improvement}
\label{sec:adapt}
\input{sections/theory/adaptivity}

\section{Numerical Examples}
\label{sec:experiments}
In this segment, we evaluate the surrogate modeling techniques mentioned in Section~\ref{sec:method} using standard academic examples and two real-world problems, each presenting its distinct challenges. Subsequently, we conduct a comparative and critical analysis of the strengths and limitations of each approach in varied scenarios. The method outlined in Section~\ref{sec:method} was implemented in Python programming language using GPflow library~\cite{GPflow2017}. This implementation is open source and can be accessed on GitHub\footnote{https://github.com/KislayaRavi/MuDaFuGP}. 
\subsection{Academic Examples}
\label{subsec:academic}
\input{sections/results/academic_problems}

\subsection{Terramechanical example}
\label{subsec:terra}
\input{sections/results/terra_mech}

\subsection{Plasma-physics example}
\label{subsec:plasma}
\input{sections/results/plasma}

\section{Conclusions}
\label{sec:conclusions}
\input{sections/conclusion}

\section*{Acknowledgments}
\input{sections/acknowledgement}

\newpage
\section*{Reference}
\bibliographystyle{unsrt}
\bibliography{mfgp}

\newpage
\appendix
\section*{Appendix}
\input{sections/appendices}
\end{document}

%% file: sections/authorlist.tex
\author{Kislaya Ravi\dag *, Vladyslav Fediukov\dag\ddag *, Felix Dietrich\dag,  Tobias Neckel\dag, Fabian Buse\ddag, Michael Bergmann\P, Hans-Joachim Bungartz\dag}

\address{\dag\ School of Computation, Information and Technology, Technische Universität München, Garching b. München, Germany}
\address{\ddag\ Institute of System Dynamics and Control, German Aerospace Center (DLR), Weßling, Germany}
\address{\P\ Max-Planck-Institut für Plasmaphysik, Garching b. München, Germany}

\address{* These authors contributed equally to this work}


%% file: sections/abstract.tex
One of the main challenges in surrogate modeling is the limited availability of data due to resource constraints associated with computationally expensive simulations. Multi-fidelity methods provide a solution by chaining models in a hierarchy with increasing fidelity, associated with lower error, but increasing cost.
In this paper, we compare different multi-fidelity methods employed in constructing Gaussian process surrogates for regression. Non-linear autoregressive methods in the existing literature are primarily confined to two-fidelity models, and we extend these methods to handle more than two levels of fidelity.
Additionally, we propose enhancements for an existing method incorporating delay terms by introducing a structured kernel. We demonstrate the performance of these methods across various academic and real-world scenarios.
Our findings reveal that multi-fidelity methods generally have a smaller prediction error for the same computational cost as compared to the single-fidelity method, although their effectiveness varies across different scenarios.

%% file: sections/intro.tex
Complex simulations are often computationally expensive and time-consuming. A surrogate model is a simpler and faster model that emulates the output of a complex model as a function of input parameters. Surrogates, also known as digital twins, have various applications in design optimization~\cite{jiang2020surrogate, mack2007surrogate}, uncertainty quantification~\cite{wang2022recent, farcaș2018nonintrusive}, real-time predictions~\cite{yondo2018review, balu2022physics}, etc. The amount of training data is one of the crucial factors governing the quality of the surrogate. Generating an extensive training data set is computationally infeasible for expensive simulations. In this work, we tackle the data availability limitation using Multi-fidelity~\cite{peherstorfer2018survey}. 

The advancement of computational capabilities significantly increased the use of numerical simulation in almost every field of application. 
This led to the development of various simulation methods offering different levels of approximation quality and computational costs.
Low-fidelity models provide estimations with reduced accuracy but demand fewer resources, whereas high-fidelity models yield precise predictions at a higher cost. 
These distinct fidelity levels arise from differences in physical modeling or simulation of the same phenomena, the linearization of complex physical processes, or the application of the same modeling approach with different discretization levels. 
The trade-off between simulation accuracy and computational cost is an ever-present challenge. 
Relying solely on high-fidelity models for applications that require multiple evaluations of the model is impractical due to their computational cost. 
Conversely, utilizing only low-fidelity models may compromise the results' accuracy. 
Multi-fidelity methods address this challenge by leveraging low-fidelity models to alleviate computational load and incorporating occasional high-fidelity evaluations to control result accuracy.

Multi-fidelity methods are widely used in applications like surrogate modeling, uncertainty quantification~\cite{giles2015multilevel, ng2012multifidelity, farcaș2023context, vinod2023optimized}, bayesian inference~\cite{ravi2023multi, lykkegaard2023multilevel, prescott2020multifidelity}, optimization~\cite{agrawal2023multi, irshadleveraging, lazin2023high, dlr135334, dlr135757, dlr190308}, etc.  
They are helpful in scenarios where resource considerations and accurate estimations are important simultaneously. 

This work focuses on using multi-fidelity models to build a Gaussian Process (GP)~\cite{williams2006gaussian} surrogate. We use GP because of its probabilistic nature and additional functionalities like Bayesian optimization and uncertainty quantification.
There are multiple ways to incorporate multi-fidelity in GP. The first method developed was the linear auto-regressive models~\cite{kennedy2000predicting, le2013multi}. The linear modeling was not sufficient for more complex problems. This led to the development of non-linear auto-regression methods~\cite{lee2019linking, perdikaris2017nonlinear}. However, the non-linear methods cannot be extended to more than two levels of fidelity and require low-fidelity evaluations during prediction. One can use Deep Gaussian Process (DGP)~\cite{cutajar2019deep} to tackle the limitations. 

This paper is organized as follows. We provide the basic theoretical background on GP, kernels, and calibration in Section~\ref{sec:theory}. Then, we review different multi-fidelity GP surrogate models in Section~\ref{sec:method}. We also suggest modifications in existing non-linear autoregressive to incorporate more than two fidelities without DGP. In the same section, we suggest using delay terms in multi-fidelity DGP. Then, we compare the performance of different methods on the academic problem and two real-world problems, namely Terramechanical problems and Plasma microturbulence simulations in Section~\ref{sec:experiments}. Finally, we end the paper with concluding remarks in Section~\ref{sec:conclusions}.

%% file: sections/theory/gp.tex
Gaussian processes (GP) are an efficient and flexible tool for probabilistic predictions~\cite{williams2006gaussian}.
They provide reliable statistical inference as well as uncertainty predictions.
A GP is a stochastic process in which each finite subset of variables forms a multivariate Gaussian distribution. GPs are defined by their covariance (kernel) function $k$ and define a probability distribution over functions, 
\begin{equation}
f(\cdot , \cdot) \sim \mathcal{G}\mathcal{P}(0, k(\cdot , \cdot)). 
\end{equation}
The mean function is usually assumed to be zero for simplicity.
In practice, the base space of the process is sampled with a finite number of $N$ points $X=\left\lbrace x_i \right\rbrace_{i=1}^N$. Then, the GP is given as a finite-dimensional, multivariable Gaussian distribution of points $y$,
\begin{equation}
y \sim \mathcal{N}(\mu, K), 
\end{equation}
where $\mu$ is a vector of means, usually zeros, and $K=(K_{ij})_{i,j=1}^N=(k(x_i,x_j))_{i,j=1}^N$ is a covariance matrix of size $N\times N$. 
Gaussian distributions can be conveniently employed in a Bayesian framework because they are closed under condition and conjugate distributions, which means marginal and conditional distributions of multivariate Gaussians are again Gaussian. 

GP is typically used in regression, also called the ``Kriging method''~\cite{krige1951statistical}.
In this case, each training point $x_i\in X$ is assigned an additional function value $y_i$, and the goal is to construct function values $y^{*}$ for unobserved test points $X^{*}=\{x^*\}$.
Then, we can construct a joint distribution  

\begin{equation}
    \left[ \matrix{  f(X) \cr f(X^*) } \right] \sim \mathcal{N}\left( \mathbf{0}, \left[ \matrix{ K(X,X) & K(X,X^*) \cr K(X^*, X) & K(X^*,X^*) } \right] \right).
\end{equation}

\begin{figure}[!]
\centering
    \begin{subfigure}{\textwidth}
    \centering
    \includegraphics[width=\textwidth]{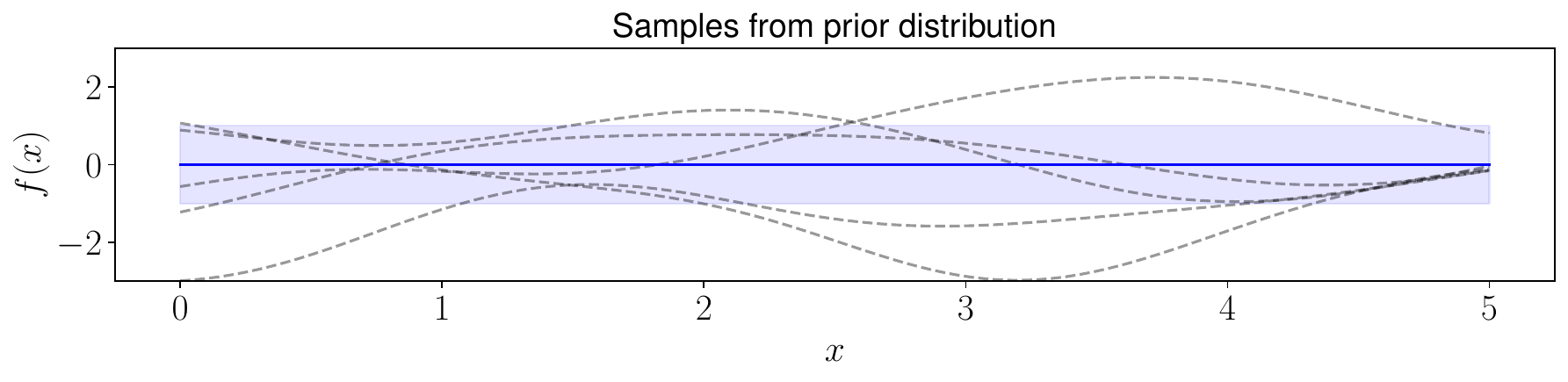}
    \caption{Visualization of Gaussian process prior.}
    \label{fig:prior_gp}
    \end{subfigure}
\newline
    \begin{subfigure}{\textwidth}
    \centering
    \includegraphics[width=\textwidth]{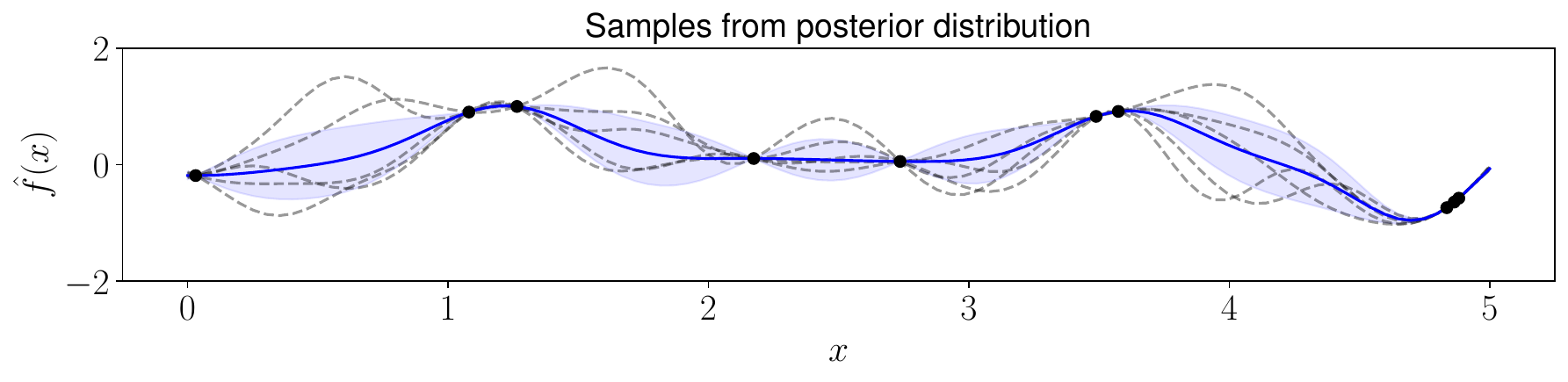}
    \caption{Visualization of Gaussian process posterior.}
    \label{fig:post_gp}
    \end{subfigure}
    \caption{A kernel function defines the family of the functions with which we are approximating the target function. In this example, we use squared-exponential or RBF kernel and visualize a) a prior distribution over the functions. b) In the posterior distribution in the second plot, newly observed training data points, depicted here in black, narrow the space of potential functions from the initial family, defined by the RBF kernel.}
    \label{fig:prior_post_gp}
\end{figure}

After we observe the training data, the posterior consists of a narrowed distribution of functions, illustrated in Figure~\ref{fig:post_gp}, initially defined by the prior distribution as observed in Figure~\ref{fig:prior_gp}.
The inference is done then using posterior mean and posterior covariance defined as

\begin{equation}
\label{eqn:posterior_mean}
    \mathbb{E}(f(X^*)) = K(X,X^*)^{T}K(X,X)^{-1}f(X),
\end{equation}
\begin{equation}
\label{eqn:posterior_cov}
    cov(f(X^*)) = K(X^*,X^*)- K(X,X^*)^{T}K(X,X)^{-1}K(X^*,X).
\end{equation}

%% file: sections/theory/kernels.tex
$K$ denotes a matrix built with the help of a \textit{kernel}, which is a function used to define the covariances between the data points.
This covariance is usually interpreted as a measure of similarity between data points, and thus, points $x$ and $x'$ that are considered similar in the input space will have target values $y$ and $y'$ that are also similar.

A valid covariance function must be positive semidefinite~\cite{williams2006gaussian}. The product and sum of two valid covariance functions generate a valid covariance function. Using this property, one can generate more sophisticated covariance functions. 

There are multiple options for covariance functions like squared-exponential, exponential, trigonometric, etc. The task of choosing the right kernel is complicated. There is no standardized algorithm to do so except for the random search and the researcher's intuition regarding modeling. 
However, some authors are making promising progress by creating interesting heuristics and algorithms~\cite{owhadi2019kernel, huwel2021automated, duvenaud2013structure, horn2019surrogates}. 
In this work, we only work with squared-exponential covariance functions, because it yields smooth, i.e. mean-square differentiable, sample paths from the resulting GP, while others, like exponential kernel, do not~\cite{williams2006gaussian}.

After choosing the kernel family, we evaluate the hyperparameters of the kernel by maximizing the log-likelihood. 
The log-likelihood $L$ as a function of observed values of data points $Y$, parameterized by the hyperparameters $\theta$ as
\begin{equation}
\log(p(y|X, \theta)) = -\frac{1}{2}y^{T}{K_y}^{-1}y - \frac{1}{2}\log|K_y| - \frac{n}{2}\log 2\pi.
\label{eq:loglikelihoodopt}
\end{equation}
We will use a gradient-based method (LBFGS~\cite{liu1989limited} with multiple starting points) to find the maximum likelihood. 

%% file: sections/theory/calibration.tex
As shown in \RefEq{eq:loglikelihoodopt}, hyperparameter optimization for Gaussian process kernels is usually performed by maximizing the marginal likelihood. This often leads to an acceptable mean squared error~\cite{williams2006gaussian}, but the posterior variance tends to be lower than the variance of the actual distribution~\cite{capone2022gaussian, capone2023sharp}. 
This problem requires an additional post-processing step known as calibration.
There are various calibration techniques, but in all cases, they aim to adjust the predicted variance towards the actual variance in the data.
Calibration is an established concept in the classification context, but in the case of regression, it is relatively new \cite{kuppers2022parametric}.
In general, calibration can be split into three categories: \textit{quantile}, \textit{distribution}, and \textit{variance} calibration.

The regression is \textit{quantile calibrated}, if
\begin{equation}
    \frac{1}{N}\sum^{N}_{n=1} \mathbf{I} \{ y_n<F^{-1}_n(p) \} \rightarrow p \textrm{ for all } p \in [0, 1] \textrm{ as } N \rightarrow \infty 
\end{equation}
for the set of data ${(x_n, y_n)}^{N}_{n=1}$, where $F_n$ stands for predicted cumulative distribution at $x_n$ and $F^{-1}_n$ the corresponding quantile function. Intuitively, $X\%$ of the confidence interval should capture the ground truth values in $X\%$ predictions~\cite{song2019distribution, kuleshov2018accurate}.
Quantile calibration is considered a global calibration, as it concentrates only on the marginal distribution without considering a distribution calibration around each exact prediction.

\textit{Distribution calibration} resembles the calibration approach in classification tasks because it is also focused on local calibration. In the classification case, instances are grouped by their predicted probability and considered calibrated only if each group matches the indicated probability.
Assume $X$ as input variables, $Y$ as a target variable, and the regression model $f: \mathbb{X} \rightarrow \mathbb{S}_{\mathbb{Y}}$ or intuitively, the regression model is mapping from input features into a space of distributions and for each test instance predicted a particular distribution. Denoting $s$ as an arbitrary distribution predicted by the regression model, the regression model $f$ is distribution calibrated if 
\begin{equation}
    p(Y=y|f_{Y|X}=s)=s(y) \textrm{ for all } y \in Y, s \in S,
\end{equation}
or intuitively, for all predicted instances with a particular distribution $s$, all instances on average with such prediction should follow this distribution $s$~\cite{song2019distribution}.

\textit{Variance calibration} aims to make the model predict its own error, i.e. for all instances where variance $\sigma^2(x)$ takes a certain value $\omega$, the squared predicted error $\mathbb{E}_{X,Y}[\mu(x)-y]$ will match this predicted variance~\cite{levi2022evaluating}.
Given $\mu(x)$ is a predicted mean and $\sigma^2(x)$ is a predicted variance, then the regression model is variance calibrated if
\begin{equation}
    \mathbb{E}_{X, Y}[(\mu(x)-y)^2|\sigma^2(x) = \omega] = \omega.
\end{equation}

The first method performs a global calibration, while the last two perform a local one. 
Global calibration is a better choice in case we need only to rescale the marginalized distribution, but in the case when variance should be rescaled only in certain places, it is better to use local calibration.
Knowing a priori the required type of rescaling is hard for real-world data, so if we do not have any prior assumptions on the real variance, we must empirically find the most suited calibration method. 

%% file: sections/theory/mf.tex
In this paper, our focus is on exploring various multi-fidelity Gaussian process surrogate modeling methods. In the upcoming sub-sections, we delve into the motivation behind each method, elucidate their formulation, and critically evaluate their limitations. Additionally, we propose enhancements for some methods within the corresponding sub-sections to address identified shortcomings.

%% file: sections/theory/ar1.tex
We start with the Auto-Regressive Model (AR1), a straightforward linear auto-regressive model introduced by Kennedy and O'Hagan~\cite{kennedy2000predicting}. In this model, we formulate the joint distribution of all fidelities, building on the fundamental concept of expressing the model for a particular fidelity as the sum of a linear scaling of the previous fidelity and an additive correction term.

We explain AR1 using a two-fidelity case. This process can be easily extended to cases with more than two fidelities. Let us consider the following random variables:
\begin{eqnarray*}
    u_l \sim \mathcal{N}(0, K_{l}) & \Rightarrow \mathbb{E}[u_l] = 0 \quad \textrm{and} \quad \mathbb{E}[u_l u_l'] = K_{l}(X, X'), \\
    u_{\delta} \sim \mathcal{N}(0, K_{\delta}) & \Rightarrow \mathbb{E}[u_{\delta}] = 0 \quad \textrm{and} \quad \mathbb{E}[u_{\delta} u_{\delta}'] = K_{\delta}(X, X').
\end{eqnarray*}

We are modeling all the surrogates and the additive correction term using GP. $u_l$ and $u_{\delta}$ represent the samples drawn from the low-fidelity and the additive correction GP.
Let $u_h$ represent a realization of the GP that approximates a high-fidelity function $f_h$. $\rho \in \mathbb{R}$ is a linear-scaling parameter. We assume an additive relationship between the consecutive fidelities so that
\begin{equation}
    u_h = \rho u_l + u_{\delta}.
    \label{eq:AR1-formulation}
\end{equation}
The expected value of the surrogate of the high-fidelity model is assumed to be zero,
\begin{equation}
    \mathbb{E}[u_h] = \mathbb{E}[\rho u_l + u_{\delta}] = \rho\mathbb{E}[u_l] + \mathbb{E}[u_{\rho}] = 0.
\end{equation}
The covariance of the surrogate of the high-fidelity model is then given by
\begin{eqnarray*}
    Cov(u_h, u_h') &= \mathbb{E}[u_h u_h'] - \underbrace{\mathbb{E}[u_h] \mathbb{E}[u_h']}_{=0} \\ 
    &= \mathbb{E}\left[\left(\rho u_l + u_{\delta}\right) \left(\rho u_l'\right) + u_{\delta}' \right] = \rho^2 \mathbb{E}[u_l u_l'] + \mathbb{E}[u_{\delta} u_{\delta}'] = \rho^2 K_l + K_{\delta}.
\end{eqnarray*}
The covariance between $u_l$ and $u_h$ is
\begin{eqnarray*}
    Cov(u_h, u_l') &= \mathbb{E}[u_h u_l'] - \underbrace{\mathbb{E}[u_h] \mathbb{E}[u_l']}_{=0} = \mathbb{E}\left[\left(\rho u_l + u_{\delta}\right) u_l' \right] \\
    &= \rho\mathbb{E}[u_l u_l'] + \mathbb{E}[u_{\delta} u_l'] = \rho K_l + \underbrace{\mathbb{E}[u_{\delta}] \mathbb{E}[u_l']}_{=0} = \rho K_l.
\end{eqnarray*}
The joint distribution of samples drawn from the low-fidelity surrogate and the high-fidelity surrogate is written
\begin{equation}
    \left[ \matrix{
    u_h \cr u_l
    }\right] \sim \mathcal{N} \left( 
    \left[ \matrix{
    0 \cr 0
    }\right], 
    \left[ \matrix{
    K_{\delta} + \rho^2 K_l & \rho K_l \cr \rho K_l & K_l
    } \right]
    \right).
\label{eq:AR1-joint-distribution}
\end{equation}
Using the maximum likelihood method, one can calculate the value of $\rho$ and the kernel hyperparameters. We can also extend the joint distribution in \RefEq{eq:AR1-joint-distribution} to incorporate more than two fidelities that will lead to covariance matrices with sparse blocks. Le Gratiet~\cite{le2013multi} took advantage of this block structure and decreased the complexity of the method from $\mathcal{O}\left((\sum_{l=1}^{L} N_l)^3\right)$ to $\sum_{l=1}^{L} \mathcal{O}( N_l^3)$. He also improved the performance using polynomial terms for scaling ($\rho$) instead of a constant term.

Nevertheless, the efficacy of the Auto-Regressive Model (AR1) is limited by the assumption of a linear dependency between the low-fidelity and high-fidelity functions. This assumption might prove inadequate in scenarios characterized by non-linear transformations. 

%% file: sections/theory/data_fusion.tex
Let us consider a two-fidelity case. A non-linear transformation on the low-fidelity function to obtain the high-fidelity function is written as
\begin{equation}
    f_h(x) = g\left(x, f_l(x)\right),
    \label{eq:nonlinear-transformation}
\end{equation}
where $g:\mathbb{R}^{d+1} \rightarrow \mathbb{R}$ is a function representing the non-linear transformation applied on the low-fidelity function to obtain the high-fidelity model.
When attempting to model both the low-fidelity and high-fidelity functions using GPs, writing the joint distribution of the surrogates, as presented in \RefEq{eq:AR1-joint-distribution}, becomes analytically challenging for a general transformation $g$. 
To overcome this difficulty, a common assumption in studies such as~\cite{perdikaris2017nonlinear, lee2019linking} is that the low-fidelity function can be evaluated at zero cost. This eliminates the need for a surrogate for the low-fidelity function. Consequently, the non-linear transformation depicted in \RefEq{eq:nonlinear-transformation} transforms into creating a surrogate in a $d+1$ dimensional space. In our work, we delve into the modeling of $g$ using a Gaussian Process (GP). This approach enables us to effectively navigate the challenges associated with the joint distribution of surrogates, providing a viable solution for incorporating non-linear transformations within the multi-fidelity surrogate modeling framework.

Let us consider the following pedagogical examples of low-fidelity and high-fidelity functions from~\cite{perdikaris2017nonlinear}. 
The high-fidelity model $f_{h}$ and the lower-fidelity model $f_{l}$ are defined through
\begin{eqnarray*}
    f_h(x) &=& (x - \sqrt{2}) \sin^2\left( 8 \pi x \right), \\
    f_l(x) &=& \sin \left( 8 \pi x \right).
\end{eqnarray*}
\begin{figure}[!t]
    \centering
    \includegraphics[scale=0.99]{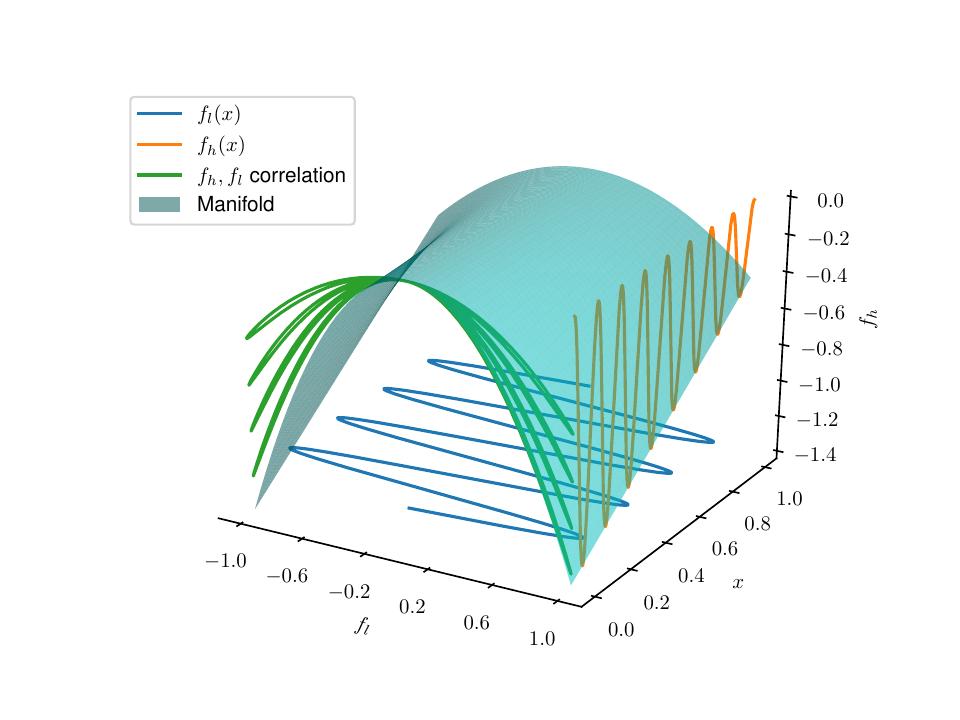}
    \caption{It is difficult to approximate the function $f_{h}(x)$ (orange) using only $x$ as an input, because it is highly oscillatory. Including the function $f_{l}(x)$ (blue) as an additional input means only the green, smooth manifold must be approximated, which can be done accurately with only a few data points. The example is adapted from \cite{perdikaris2017nonlinear}.}
    \label{fig:nonlinear-manifold}
\end{figure}
%
Figure~\ref{fig:nonlinear-manifold} illustrates the relationship between $f_l$ and $f_h$. Modeling the transformation $g$ as described in \RefEq{eq:nonlinear-transformation} is synonymous with learning the manifold depicted in Figure~\ref{fig:nonlinear-manifold}.
Crucially, this manifold lacks oscillations, signifying that learning it necessitates relatively fewer high-fidelity function evaluations. This observation underscores the advantage of multi-fidelity modeling in this example, as it allows for a learning process by leveraging the complicated feature to be captured in the low-fidelity function, thereby reducing the requirement of high-fidelity function evaluations.

A refinement of the method involves incorporating a well-structured kernel into the GP model, as suggested by Peridikaris et al.~\cite{perdikaris2017nonlinear}. The proposed kernel structure is given by:
\begin{equation}
    k(x, x') = k_{\rho}(x, x' ; \theta_{\rho}) k_f (f_l(x), f_l(x'); \theta_f) + k_{\delta}(x, x' ; \theta_{\delta}).
    \label{eq:NARGP-kernel}
\end{equation}
Here, $\theta_{\rho}$, $\theta_f$, and $\theta_{\delta}$ denote the kernel hyperparameters, which are determined by maximizing the likelihood, as discussed in the previous section. This kernel structure closely mirrors the Auto-Regressive Model (AR1) formulation described in \RefEq{eq:AR1-formulation}, assigning each kernel to an individual part of the transformation. Specifically, $k_{\rho}$ models the scaling term, $k_f$ is responsible for transforming the low-fidelity model, and $k_{\delta}$ handles the additive correction term. The formulation above gives the method its name: Non-linear Auto-Regressive Gaussian Process (NARGP).

Introducing this well-defined prior aids the model in accurately fitting the underlying relationships. A drawback of this formulation is the increase in the number of hyperparameters, which necessitates careful consideration during the optimization process. Despite this challenge, the structured nature of the kernel contributes to a more informed and effective surrogate modeling approach.

We follow Lee et al.~\cite{lee2019linking} and assume sufficient smoothness of the high-fidelity function $f_h$. We can then use Taylor expansion to approximate the value of $f_h$ close to $x$ by
\begin{equation}
    f_h(x + \Delta x) = f_h(x) + \sum_{i=1}^{d} \frac{\partial }{\partial x_i}f_h(x) \Delta x_i + \mathcal{O}(\Delta x^2).
    \label{eq:taylor-hf}
\end{equation}
Substituting \RefEq{eq:nonlinear-transformation} into \RefEq{eq:taylor-hf}, we obtain
\begin{eqnarray*}
    f_h(x + \Delta x) &=& g(x, f_l(x)) + \sum_{i=1}^{d} \left(\frac{\partial g}{\partial x_i}g(x, f_l(x)) \right)\Delta x_i + \mathcal{O}(\Delta x^2) \\
    &=& g(x, f_l(x)) + \sum_{i=1}^{d} \left(\frac{\partial g}{\partial x_i} + \frac{\partial g}{\partial f_l} \frac{\partial f_l}{\partial x_i} \right)\Delta x_i + \mathcal{O}(\Delta x^2).
\end{eqnarray*}
The Taylor expansion converges if
\begin{enumerate}
    \item $ \left\Vert\frac{\partial g(\cdots)}{\partial x} \right\Vert \leq c_g$ for $c_g \in \mathbb{R}^+$ 
    \item $\left\Vert \frac{\partial f_l(\cdots)}{\partial x} \right\Vert \leq c_l$ for $c_l \in \mathbb{R}^+$
\end{enumerate}
The Taylor expansion also motivates us to incorporate the derivative of low-fidelity in \RefEq{eq:nonlinear-transformation}.
Lee et al.~\cite{lee2019linking} suggest writing the non-linear transformation as
\begin{equation}
    f_h(x) = g_d \left(x, f_l(x), \frac{\partial f_l}{\partial x} \right).
    \label{eq:nonlinear-transformation-der}
\end{equation}
In many legacy simulators, the gradients are unavailable~\cite{cranmer2020frontier}. In cases where gradients are unavailable, one can resort to finite difference techniques to approximate the derivative.
 Lee et al.~\cite{lee2019linking} suggest replacing the derivative in \RefEq{eq:nonlinear-transformation-der} with the delay term to accommodate such scenarios. This delay term evaluates the low-fidelity function after a small time step $\tau \in \mathbb{R}$. For a higher-dimensional input variable, we add a delay along each dimension, defining $\tau_i$ as a vector of zeros with $\tau$ at the $i^{\textrm{th}}$ position. 
The transformed formulation is now expressed as
\begin{equation}
    f_h(x) = g_d \left(x, f_l(x), f_l(x+\tau_1), f_l(x+\tau_2), \ldots, f_l(x+\tau_d)  \right).
    \label{eq:nonlinear-transformation-delay}
\end{equation}
This method is called the Gaussian Process with Delay Fusion (GPDF) proposed in~\cite{lee2019linking}. 
Additionally, we propose an enhancement to this method by modifying the kernel to mimic the Taylor expansion using a structured composite kernel similar to \RefEq{eq:NARGP-kernel}
\begin{equation}
    k(x, x') = k_{\rho}(x, x' ; \theta_{\rho}) k_f ((f_l(x), f_l(x+\tau)), (f_l(x'), f_l(x'+\tau)); \theta_f) + k_{\delta}(x, x' ; \theta_{\delta}).
    \label{eq:GPDFC-kernel}
\end{equation}
This improved method is called the Gaussian Process with Delay Fusion and Composite kernel (GPDFC).

The formulations of NARGP, GPDF, and GPDFC come with certain limitations that need to be addressed for broader applicability:
\begin{enumerate}
    \item \textbf{Nested Evaluation Requirements:}
    \begin{itemize}
        \item NARGP: Requires low-fidelity model evaluations during training and prediction at precisely the same parameters as the high-fidelity model.
        \item GPDF and GPDFC: Share similar constraints and involve additional low-fidelity function evaluations at delay points.
    \end{itemize}
    These limitations were initially disregarded under the assumption of infinitely cheap low-fidelity function evaluations. However, in practice, even low-fidelity function evaluations have nonzero cost, and so these constraints can also become significant. In some cases, we are simply given data collected from the low and high-fidelity models. Gathering any further data might not be possible, and for such cases, ensuring nested training data points may also not be possible. Moreover, the possible absence of low-fidelity data at the prediction parameter point also serves as a hurdle in implementing the methods mentioned in data-driven multi-fidelity problems \cite{cutajar2019deep}.
    \item \textbf{Limited to Two Fidelity Cases:}
    Modern NARGP, GPDF, and GPDFC formulations are tailored for scenarios involving only two fidelity models, restricting their application in situations with more than two fidelities.
    \item \textbf{Overfitting:} As shown in \cite{cutajar2019deep}, NARGP is prone to overfitting, which leads to poor generalization.  
\end{enumerate}

\begin{figure}[!t]
    \begin{subfigure}[t]{0.43\textwidth}
    \centering
    \includegraphics[width=0.9\textwidth]{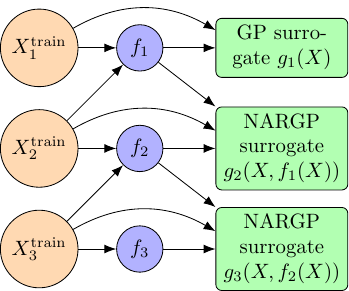}
    \caption{Training}
    \label{fig:nargp-general-train}
    \end{subfigure}
    \begin{subfigure}[t]{0.52\textwidth}
    \centering
    \includegraphics[width=0.9\textwidth]{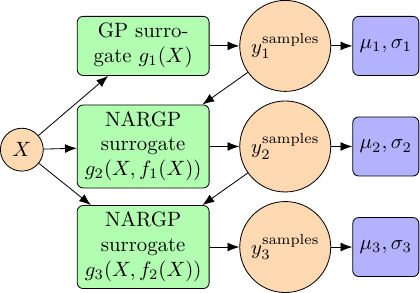}
    \caption{Prediction}
    \label{fig:nargp-general-predict}
    \end{subfigure}
    \caption{Flowchart showing the training and prediction steps of NARGP surrogate for a three-fidelity scenario. $f_1$, $f_2$, and $f_3$ represent the function at each fidelity. $g_1$, $g_2$, and $g_3$ represent the surrogate of the corresponding fidelity. Training the NARGP model for a particular level involves evaluating the model of that fidelity level and the previous fidelity level. We draw samples to make predictions for a particular level, which involves drawing samples from the surrogate of the previous level and feeding them as input to the surrogate of that level. We can then use the samples to obtain the required statistical moments of the predictions by marginalizing the samples.}
\end{figure}

To overcome the limitation of application to a two-fidelity case and extend the framework to accommodate multiple fidelity levels, we can explore two scenarios:
\begin{itemize}
    \item \textbf{Nested training data:} In the case of nested training data, where ($X_{\ell + 1}^{\textrm{train}} \subset X_{\ell}^{\textrm{train}}$), we propose training a Gaussian Process surrogate for the lowest fidelity and multi-fidelity Gaussian Process surrogates for higher levels. We train each multi-fidelity surrogate using the previous level, as shown in Figure \ref{fig:nargp-general-train} for three fidelity NARGP surrogates. These layers are then cascaded, forming a stacked surrogate for multiple fidelities. During prediction, we draw samples from the surrogate of the previous level to obtain prediction samples for the next level and marginalize it to derive the posterior prediction for the given level.  Let us represent $y_{\ell}$ as posterior prediction at parameter $X$. We can evaluate the posterior distribution of $y_{\ell+1}$ as
    \begin{equation}
        p(y_{\ell + 1}|X) = \int p(y_{\ell + 1}|y_{\ell},X) p(y_{\ell}|X) dy_{\ell}.
    \end{equation}
    One can write the marginalization step after cascading as
    \begin{eqnarray*}
        p(y_{\ell + 1}|X) = \int_{y_{\ell}} \int _{y_{\ell-1}} \cdots \int_{y_1} & p(y_{\ell + 1}|y_{\ell},X)p(y_{\ell}|y_{\ell-1}X) \\ &\ldots p(y_2|y_1, X) p(y_1|X) dy_{\ell} dy_{\ell-1} \ldots dy_1.
    \end{eqnarray*}
        
    We can perform this integration using the Monte Carlo method. The prediction steps for three fidelity stacked NARGP surrogates are shown in Figure \ref{fig:nargp-general-predict}.
    
    Every layer in the stacked architecture is a GP. So, the samples drawn from GP will stay close to the posterior mean when the posterior predicted variance is small. If we ensure that there are enough data points in each layer such that the maximum of the posterior variance across the parameter space is smaller than a cut-off limit then we can use the posterior predicted mean of the previous layer as input for the next layer. In this way, we can bypass the Monte Carlo step. An efficient way to ensure small posterior variance is by using the adaptivity algorithm, which we will discuss later.
    
    \item \textbf{Non-nested training data:} In situations with non-nested training data, an alternative is to employ the Deep Gaussian Process (DGP). 
\end{itemize}

%% file: sections/theory/deepGP.tex
Drawing inspiration from the stacked architecture of multilayer neural networks, Deep Gaussian Processes (DGP) extend a single GP by composing multiple GP with each other. Each node within DGP represents a Gaussian process. The fundamental concept of deep architectures lies in their ability to model complex functions through combinations of simpler ones, with the output of one layer serving as the input for the next~\cite{damianou2013deep}. The primary advantage of Deep Gaussian Processes (DGP) over a single Gaussian Process (GP) lies in the distribution of feature learning across different layers. In a DGP, each layer is dedicated to learning distinct features of the model. This is in contrast to a single GP, where all features are consolidated into one, potentially leading to a complex and intricate model, especially when using traditional kernels.

DGP can effectively capture various aspects or representations of the underlying system by distributing the learning process across multiple layers. DGP emerges as a suitable candidate for modeling multi-fidelity scenarios. With each fidelity level, DGP gradually incorporates new features, enhancing the model's capabilities as it approaches high-fidelity representations. This progressive refinement allows DGP to effectively capture the intricacies of complex systems.

We model the transformation similar to \RefEq{eq:nonlinear-transformation} with one difference. We replace the actual function evaluation of the previous layer with the previous GP layer 
\begin{equation}
   f_{\ell}(X) = g_{\ell}(X, g_{\ell -1}(\cdots)).
\end{equation}

The analytic form of the posterior of the GP surrogate $g_{\ell}$ is intractable because one of its parameters ($g_{\ell -1}$) is sampled from a posterior distribution. To sample from the posterior of $g_{\ell}$, we use variational inference for each layer of DGP~\cite{titsias2010bayesian}. We use $M_{\ell}$ inducing points to approximate each layer representing a fidelity level. Typically, the number of inducing points is significantly smaller than the number of training data points  ($M_{\ell} \ll N_{\ell}$). This decreases the computational requirements of GP training and prediction steps~\cite{bach2005predictive, bauer2016understanding, quinonero2005unifying}. This approach is also known as the Sparse Variational Gaussian Process (SVGP)~\cite{burt2019rates}.
We discuss the modeling approach using DGP for multi-fidelity surrogate modeling in~\cite{cutajar2019deep, salimbeni2017doubly}.

\begin{figure}[!t]
\begin{subfigure}[t]{0.46\textwidth}
\centering
\includegraphics[width=0.9\textwidth]{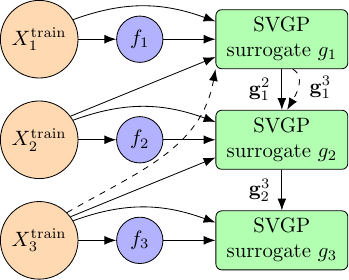}
\caption{Training}
\label{fig:dgp-train}
\end{subfigure}
\begin{subfigure}[t]{0.49\textwidth}
\centering
\includegraphics[width=\textwidth]{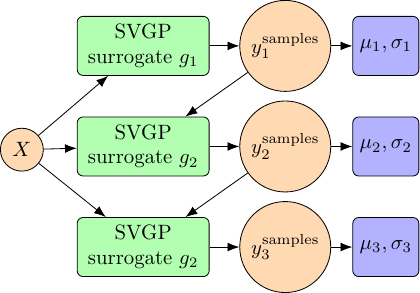}
\caption{Prediction}
\label{fig:dgp-predict}
\end{subfigure}
\caption{Flowchart showing the training and prediction steps of Deep Gaussian process surrogate for three fidelity scenarios. $f_1$, $f_2$, and $f_3$ represent the function at each fidelity. $g_1$, $g_2$, and $g_3$ represent the surrogate of the corresponding fidelity. Training the DGP model for a particular level involves evaluating the model of that fidelity level and the surrogate of the previous fidelity level, which further involves evaluating the one-level lower surrogate. This creates a cascading effect until we reach the lowest fidelity surrogate. The prediction process is similar to the full GP, which involves drawing and marginalizing samples.}
\end{figure}
Let us assume that we have a set of locations of the training parameters as 
$$\mathbf{X} = \{ X_{1}^{\textrm{train}}, X_{2}^{\textrm{train}},\ldots, X_{L}^{\textrm{train}}\}$$
and the corresponding function evaluations
$$\mathbf{Y} = \{\mathbf{y}_1, \mathbf{y}_2, \ldots, \mathbf{y}_L \} = \{ f_1 \left( X_{1}^{\textrm{train}} \right), f_2 \left( X_{2}^{\textrm{train}} \right), \ldots, f_L \left( X_{L}^{\textrm{train}} \right) \}. $$
Note that we do not impose the limitation that the location of the training parameters should be nested. Let $\mathbf{g}_{\ell}^{t}$ represent the samples drawn from the $\ell^{\textrm{th}}$ layer at the training parameter $X_{t}^{\textrm{train}}$. We assume that $M_{\ell}$ inducing points are sufficient to represent the distribution at layer $\ell$. We represent the location of the inducing point by $\mathbf{Z}_{\ell}$. Let $\mathbf{u}_{\ell}$ be the sample drawn from the $\ell^{\textrm{th}}$ layer at the inducing points $\mathbf{Z}_{\ell}$. We represent the kernel function and the mean function of the prior at layer $\ell$ using $k_{\ell}$ and $m_{\ell}$ respectively.
The unnormalized posterior distribution at $\ell^{\textrm{th}}$ layer is
\begin{eqnarray}
    p(\mathbf{y}_{\ell}, \mathbf{g}_{\ell}^{\ell}, \mathbf{u}_{\ell}|  \mathbf{X}_{\ell}, \mathbf{g}_{\ell-1}^{\ell},\mathbf{Z}_{\ell}) = \underbrace{p(\mathbf{y}_{\ell}|\mathbf{g}_{\ell}^{\ell}, \mathbf{X}_{\ell})}_{\textrm{likelihood}} \underbrace{p(\mathbf{g}_{\ell}^{\ell}|\mathbf{u}_{\ell}, \mathbf{g}_{\ell-1}^{\ell}, \mathbf{X}_{\ell}, \mathbf{Z}_{\ell}) p(\mathbf{u}_{\ell}|\mathbf{Z}_{\ell})}_{\textrm{DGP prior}}.
    \label{eq:exact-posterior-dgp}
\end{eqnarray}
    
Figure~\ref{fig:dgp-train} shows the training process for a three-fidelity scenario. The calculation of posterior at any layer needs samples from the previous layer at the training points, which in turn requires further evaluation from previous layers. For example, the calculation of the posterior of the third layer needs to draw samples from the second layer at $X_{3}^{\textrm{train}}$, which also requires samples from the first layer at $X_{3}^{\textrm{train}}$. We represent this hidden calculation with a dashed arrow in Figure~\ref{fig:dgp-train}.

One can combine all the layers by multiplying the posterior from individual layers from \RefEq{eq:exact-posterior-dgp} to obtain the joint posterior distribution for all layer $p(\mathbf{y}, \{\mathbf{g}_{\ell}^{t}, u_{\ell} \})$. Analytical inference from the posterior is intractable. To approximate the distribution, we assume that the prior $q(\mathbf{u}_{\ell})$ of $\mathbf{u}_l$ is a Gaussian distribution with mean $\tilde{\mathbf{m}}_{\ell}$ and variance $\tilde{\mathbf{S}}_{\ell}$. Now, the approximated DGP prior is
\begin{equation}
    q(\mathbf{g}_{\ell}^{\ell}, \mathbf{u}_{\ell}| \mathbf{X}_{\ell}, \mathbf{g}_{\ell-1}^{\ell},\mathbf{Z}_{\ell}) = p(\mathbf{g}_{\ell}^{\ell}|\mathbf{u}_{\ell}, \mathbf{g}_{\ell-1}^{\ell}, \mathbf{X}_{\ell}, \mathbf{Z}_{\ell}) q(\mathbf{u}_{\ell}).
    \label{eq:approx-dgp-prior}
\end{equation}
If $\mathbf{u}_{\ell}$ is marginalized out of the \RefEq{eq:approx-dgp-prior}, then the distribution becomes a Gaussian distribution with the following mean and variance:
\begin{eqnarray}
    \mu_{\ell}(x) &=& m_{\ell}(x) + \alpha(x)^T (\tilde{\mathbf{m}}_{\ell} - m_{\ell}(x)), \\
    \Sigma_{\ell}(x, x') &=& k_{\ell}(x, x') - \alpha(x)^T (k_{\ell}(\mathbf{Z}_{\ell}, \mathbf{Z}_{\ell})- \tilde{\mathbf{S}}_{\ell}) \alpha(x').
\end{eqnarray}
We can use the before approximating the posterior at each layer and then extend this to all the layers of DGP to obtain the approximate posterior distribution $q(\mathbf{y}, \{\mathbf{g}_{\ell}^{\ell}, \mathbf{u}_{\ell} \}_{\ell=1}^{L})$.
We want to approximate the exact posterior distribution in \RefEq{eq:exact-posterior-dgp} by an approximate distribution  $q(\mathbf{y}, \{\mathbf{g}_{\ell}^{\ell}, \mathbf{u}_{\ell} \}_{\ell=1}^{L})$. We train the DGP by minimizing the Kullback-Leibler (KL) divergence $\textrm{KL}[q||p]$ between the variational posterior distribution $q$ and the true posterior distribution $p$. An exact evaluation of the KL divergence is not feasible. So, we convert the minimization of the KL divergence to the maximization of Evidence Lower Bound (ELBO) ($\mathcal{L}$) whose evaluation is feasible. ELBO is evaluated as
\begin{equation}
    \mathcal{L} = \sum_{\ell=1}^{L} \mathbb{E}_{q(\mathbf{g}_{\ell}^{\ell}, \mathbf{u}_{\ell})} [ \log \underbrace{p \left(\mathbf{y}_{\ell} | \mathbf{g}_{\ell}^{\ell} \right)}_{\textrm{likelihood}} ] + KL[q(\mathbf{u_{\ell}})||p(\mathbf{u}_{\ell}|\mathbf{Z}_{\ell}))].
\end{equation}
One can refer to~\cite{salimbeni2017doubly} for detailed proof. We optimize for the variational parameters ($\mathbf{m}_{\ell}$, $\mathbf{S}_{\ell}$, $\mathbf{Z}_{\ell}$) and the hyperparameters of the kernels ($k_{\ell}$) using the Adam optimization method.

The prediction process in Deep Gaussian Processes (DGP) follows a recursive transfer of results from the previous layer into the next layer. We draw samples at each layer from the approximate distribution defined by \RefEq{eq:approx-dgp-prior}. We fuse the prediction parameter $X^*$ in intermediate layers with the sample drawn from the previous layer. We then use the fused parameter to draw a sample from that layer. We repeat this process until we have $S$ samples for each layer. We use the collected samples to compute the posterior mean and variance through a Monte Carlo approximation. Figure~\ref{fig:dgp-predict} visually illustrates these prediction steps for three fidelity models, depicting the recursive transfer and sampling process at each layer.

One can improve the surrogate by choosing a more structured kernel. Cutajar et al.~\cite{cutajar2019deep} suggest using the kernel defined in \RefEq{eq:NARGP-kernel}. We call this method the Non-linear Auto-Regressive Deep Gaussian Process (NARDGP). We suggest using the delay term to further add useful information. We call the method Deep Gaussian Process with Delay Fusion (DGPDF). We further improve upon the suggested method using a structured kernel as mentioned in \RefEq{eq:GPDFC-kernel}. We named the resulting method Deep Gaussian Process with Delay Fusion and Composite kernel (DGPDFC).

Methods discussed in Section~\ref{sec:nonliner-ar} and Section~\ref{sec:dgp} have similar ideas but different approaches to approximating the posterior distribution. In future discussions, we will sometimes mention the methods in Section~\ref{sec:nonliner-ar} as full GP to differentiate it from DGP because they do not involve sparse variational approximation of the posterior.

%% file: sections/theory/adaptivity.tex
In many cases, the available training data may not provide sufficient information to accurately model predictions, resulting in poor performance. One approach to enhance the model is adding more data points to the training set. This comes with the added cost of evaluating the corresponding function at the newly added training data locations. Therefore, judiciously choosing where to add points is crucial. The goal is to ensure that we capture the essential features of the target functions using as few training data points as possible.

In this section, we discuss an adaptive algorithm~\cite{villemonteix2009informational}. It is designed to enhance training datasets applicable to any of the methods mentioned earlier, all of which utilize combinations of Gaussian processes. This adaptive algorithm improves the overall surrogate prediction by focusing on one Gaussian process at a time.
Suppose we are training a GP using $n$ data points $\mathcal{D}_n = \{X_1, X_2, \ldots, X_n \}$. 
Let $y^{n+1}$ be the value of a sample from GP at $X^{n+1}$.  We select the next point $X_{n+1}$ where we gain maximum information which is defined as follows
\begin{equation}
    \mathcal{I}(X_{n+1}|\mathcal{D}_n) = H(y^*|\mathcal{D}_n) - H(y^*|\mathcal{D}_n \cup X_{n+1}),
\end{equation}
where $H(y^{n+1}|\mathcal{D}_n)$ is the conditional entropy. We suggest a greedy algorithm to selecting the next point $X_{n+1}$ by solving the optimization problem
\begin{equation}
    X_{n+1} = \argmax_{X_{n+1}} \mathcal{I}(X_{n+1}|\mathcal{D}_n).
    \label{eq:adapt-information}
\end{equation}
Upon simplification, it can be shown that for GPs, the gain in information is directly proportional to the logarithm of the posterior predicted standard deviation. Thus, the optimization problem stated in \RefEq{eq:adapt-information} can be converted to
\begin{equation}
    X_{n+1} = \argmax_{X^*} \sigma_p(X^*|\mathcal{D}_n),
    \label{eq:adapt-var}
\end{equation}
where $\sigma_p$ is the predicted standard deviation.
We can keep adding new points for a particular fidelity level until the maximum posterior standard deviation is below a certain specified threshold level. After that, we can move to the next level until we cover all the fidelity levels.

There are certain drawbacks to this method. Firstly, this method adds one point at a time, limiting its overall speed. Secondly, the suggested points are always towards the boundary in high-dimensional space. 

Various other approaches are proposed in the literature depending on alternative metrics to iteratively enhance the Gaussian Process surrogate~\cite{liu2018survey}. One can improve the posterior standard deviation adaptivity metric by adjusting it via some error information as discussed in~\cite{lin2004efficient}. However, this requires function evaluation at additional validation points, making it infeasible for computationally expensive high-fidelity functions. Query-by-committee (QBC)~\cite{seung1992query, freund1992information} is also computationally infeasible due to the need to train multiple models. Gradient-based adaptive approach~\cite{rumpfkeil2011dynamic} utilizes gradient information, which may be unavailable in many cases especially involving legacy solvers.
Moreover, gradient approximation using the finite difference method is numerically unstable and computationally expensive for high-fidelity models~\cite{gray2014automatic}. 
There are still many other adaptivity algorithms~\cite{liu2018survey,busby2009hierarchical, xu2020gaussian}. We will leave the discussion and the effects of those algorithms on multi-fidelity GP as future works.



%% file: sections/results/academic_problems.tex
Before applying the problem to real-world scenarios, we test different multi-fidelity surrogate modeling methods on some academic problems shown in Table~\ref{tab:academic-problems}, each posing a different modeling challenge. 
Researchers addressed the benchmarking of the surrogate models and there exists a variety of them~\cite{bliek2021expobench}. Still, our low-fidelity function for all the test cases is the same because we want to test the capability of different methods to learn the relationship between low and high-fidelity models. 
We deal with two fidelity cases and train the multi-fidelity GP surrogate using fifty low-fidelity and eight high-fidelity function evaluations. The high number of low-fidelity train data ensures low predictive variance for the low-fidelity Gaussian process. We use twenty-five and eight induction points in each level for all the cases involving deep Gaussian processes. Additionally, we also train a GP on high-fidelity data without multi-fidelity features. We then compare the multi-fidelity methods to the single-fidelity GP surrogate.

\begin{figure}[!]
    \centering
    \begin{center}
        \includegraphics[width=0.7\textwidth]{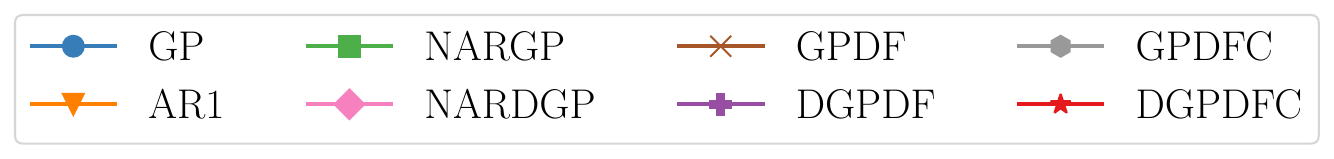}
    \end{center}
    \begin{subfigure}{.48\textwidth}
        \centering
        \includegraphics[width=\linewidth, height=0.18\textheight]{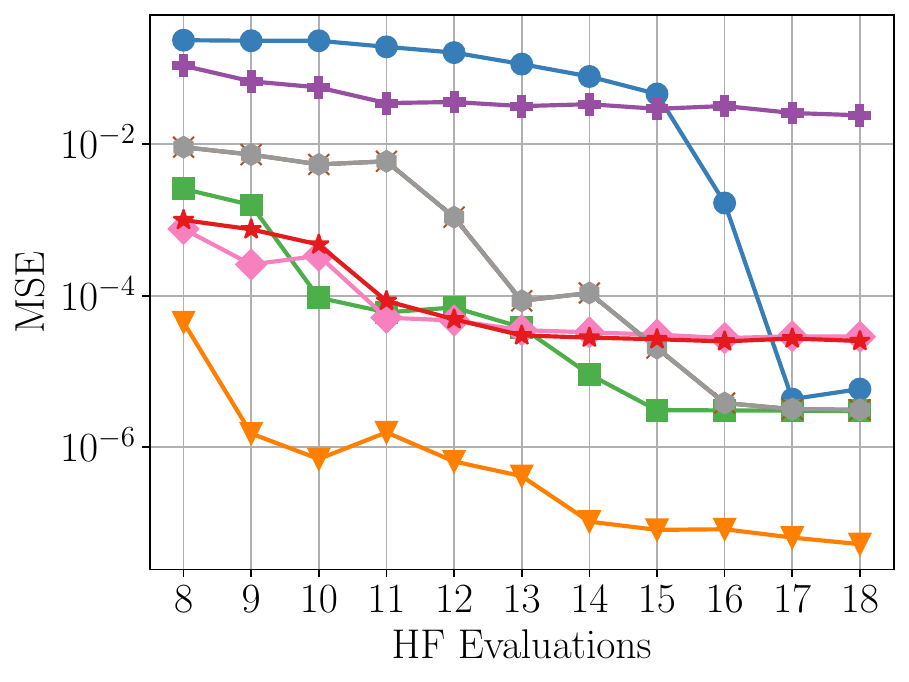}
        \caption{Linear transformation}
        \label{fig:linear-error-evolution}
    \end{subfigure}
    \begin{subfigure}{.48\textwidth}
        \centering
        \includegraphics[width=\linewidth, height=0.18\textheight]{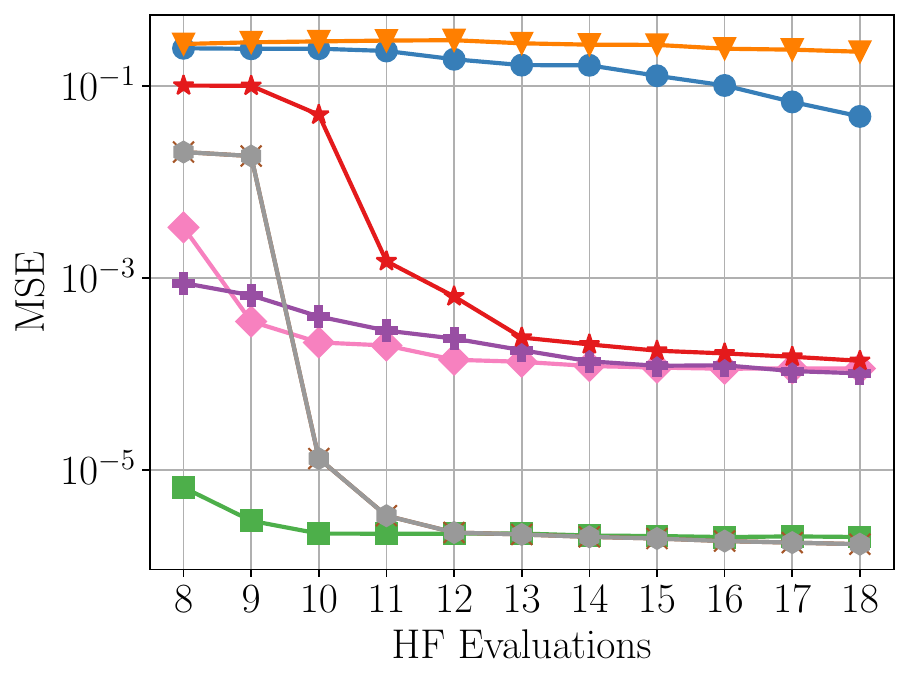}
        \caption{Non-linear transformation}
        \label{fig:nonlinear-error-evolution}
    \end{subfigure}
    \begin{subfigure}{.48\textwidth}
        \centering
        \includegraphics[width=\linewidth, height=0.18\textheight]{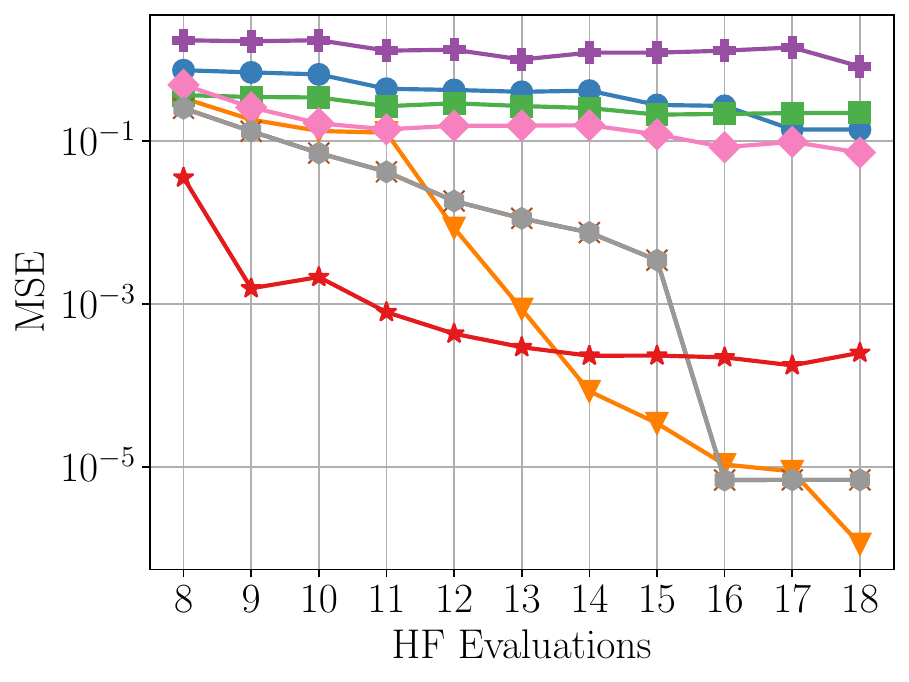}
        \caption{Phase-shifted oscillation}
        \label{fig:phase-shift-error-evolution}
    \end{subfigure}
    \caption{Average MSE evolution with respect to the number of high-fidelity function evaluations for different academic problems discussed in Section \ref{subsec:academic}}
    \label{fig:error-evolution}
\end{figure}

After this initial training, we run the adaptivity algorithm for ten steps to add ten additional high-fidelity data points. To ensure the robustness of the adaptivity algorithm, we perform multiple runs with distinct seed values. We take the average of the mean square error (MSE) and plot it against the corresponding number of high-fidelity function evaluations shown in Figure~\ref{fig:error-evolution}. The resulting plot visualizes the rate at which the respective algorithm convergences to the target high-fidelity curve. We also summarize the average MSE in Table~\ref{tab:mse_academic}.

\begin{table}[!t]
\caption{Benchmark problems for the academic examples.}
 \begin{adjustbox}{width=\columnwidth,center}
\begin{tabular}{| l | c | c |}
\hline
Benchmark name            &\begin{tabular}[c]{@{}c@{}}Low-fidelity function ($f_l$)\end{tabular}&\begin{tabular}[c]{@{}c@{}}High-fidelity function ($f_h$)\end{tabular} \\ \hline \hline
Linear transformation&& $0.8 \sin{\left(8 \pi x\right)} + 0.3 \sin{\left(2 \pi x\right)}$ \\ \cline{1-1} \cline{3-3} 
Non-linear transformation&$\sin{\left(8 \pi x\right)}$ & $\sin^2{\left(8 \pi x\right)}$ \\ \cline{1-1} \cline{3-3} 
Phase-shifted oscillation&&$\sin^2{\left(8 \pi x + \frac{\pi}{10} \right)} + \cos{\left(4 \pi x \right)}$\\ \hline
\end{tabular}
\end{adjustbox}
\label{tab:academic-problems}
\end{table}

\begin{figure}[!]
    \centering
    \begin{center}
        \includegraphics[width=0.25\linewidth]{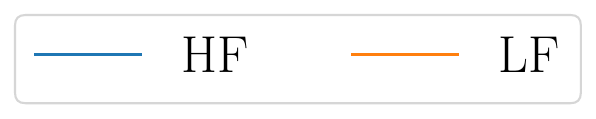}
    \end{center}
    \begin{subfigure}{.4\textwidth}
            \centering
            \includegraphics[scale=0.36]{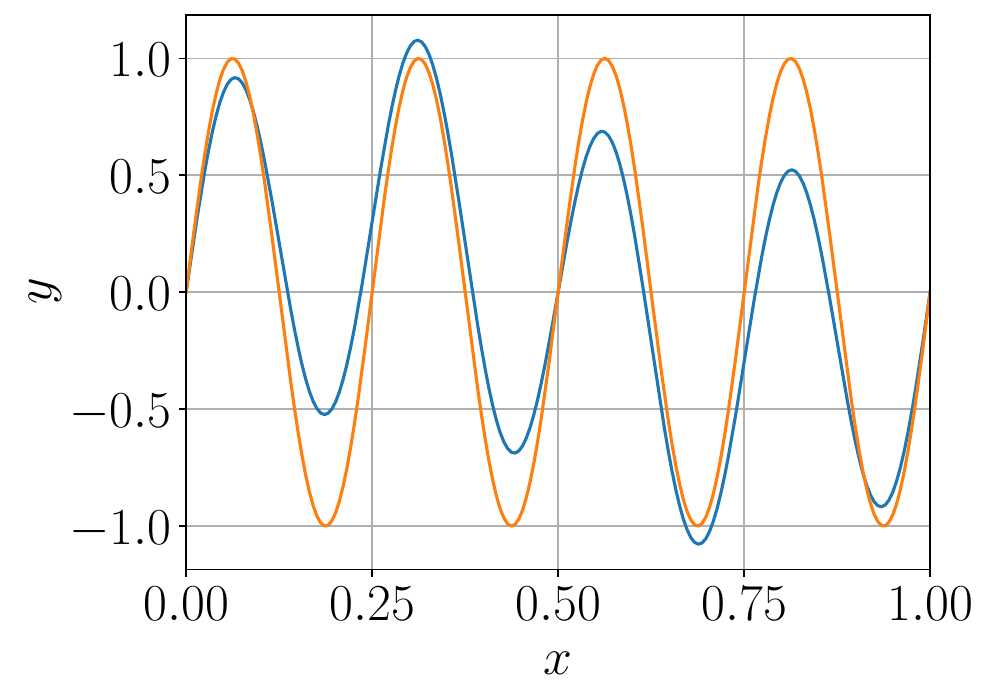}
            \caption{Linear transformation}
            \label{fig:linear-actual}
        \end{subfigure}
        \quad
    \begin{subfigure}{.4\textwidth}
            \centering
            \includegraphics[scale=0.36]{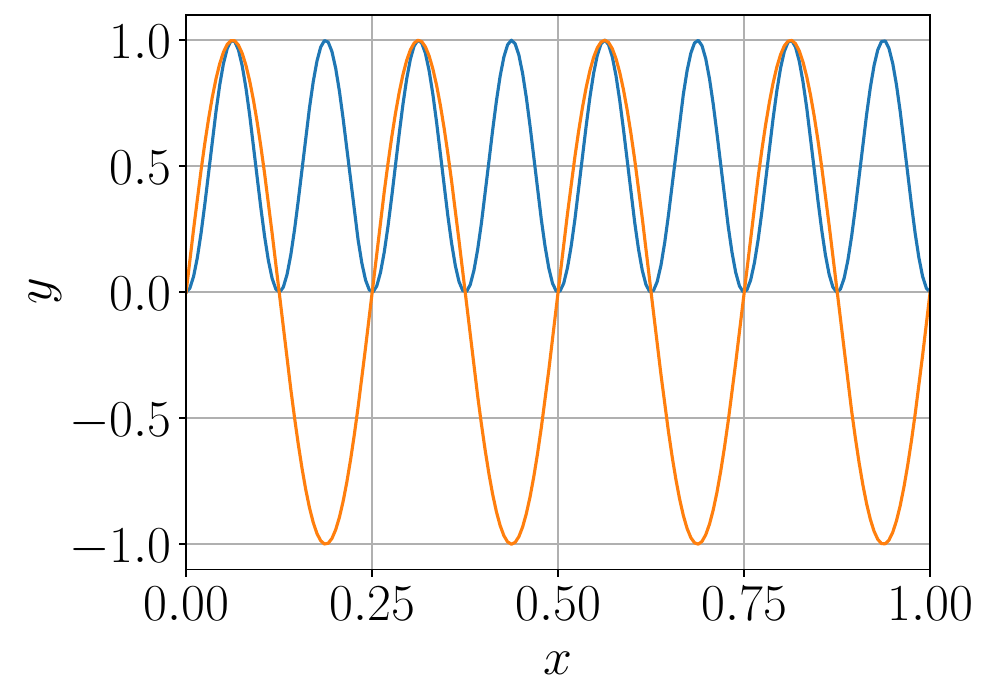}
            \caption{Non-linear transformation}
            \label{fig:nonlinear-actual}
    \end{subfigure} 
    \newline
    \begin{subfigure}{.4\textwidth}
            \centering
            \includegraphics[scale=0.36]{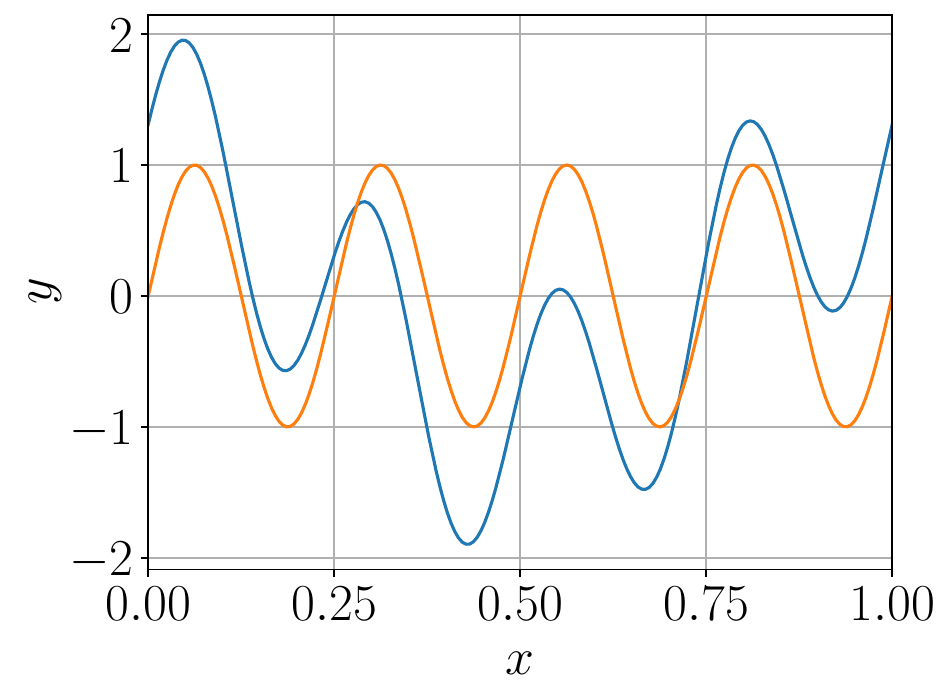}
            \caption{Phase-shifted oscillation}
            \label{fig:phase-shift-actual}
    \end{subfigure}
    \caption{Visualization of different academic problems. The low and high-fidelity functions are visualized in each sub-plot by orange and blue curves, respectively.}
    \label{fig:academic-transformations}
\end{figure}

\begin{table}[!htpb]
  \caption{Average value of MSE of each method in three different transformation scenarios, namely linear scaling, non-linear, and phase-shifted transformation as shown in Table \ref{tab:academic-problems}, w.r.t. number of high-fidelity data points used in training.}
  \begin{adjustbox}{width=\columnwidth,center}
  \begin{tabular}{*{7}{l}}
    \toprule
    & \multicolumn{6}{c}{Type of transformation} \\
    \cmidrule(lr){2-7}
    & \multicolumn{2}{c}{Linear scaling} & \multicolumn{2}{c}{Non-linear transformation} & \multicolumn{2}{c}{Phase-shifted oscillation}  \\
    \cmidrule(lr){2-3}
    \cmidrule(lr){4-5}
    \cmidrule(lr){6-7}
    MF method & 8 HF points & 18 HF points & 8 HF points & 18 HF points & 8 HF points & 18 HF points \\
    \midrule
    GP & $2.36 \times 10^{-1}$ & $5.85 \times 10^{-6}$ & $2.45 \times 10^{-1}$ & $4.79 \times 10^{-2}$ & $7.32 \times 10^{-1}$ & $1.37 \times 10^{-1}$ \\
    AR1 & $\mathbf{4.46 \times 10^{-5}}$ & $\mathbf{5.24 \times 10^{-8}}$ & $3.73 \times 10^{-1}$ & $2.71 \times 10^{-1}$ & $2.25 \times 10^{-1}$ & $\mathbf{1.14 \times 10^{-6}}$ \\
    NARGP & $2.57 \times 10^{-3}$ & $3.02 \times 10^{-6}$ & $\mathbf{6.57 \times 10^{-6}}$ & $2.00 \times 10^{-6}$ & $3.62 \times 10^{-1}$ & $2.19 \times 10^{-1}$ \\ 
    GPDF & $9.15 \times 10^{-3}$ & $3.11 \times 10^{-6}$ & $2.04 \times 10^{-2}$ & $\mathbf{1.68 \times 10^{-6}}$ & $2.50 \times 10^{-1}$ & $6.99 \times 10^{-6}$ \\
    GPDFC & $9.15 \times 10^{-3}$ & $3.11 \times 10^{-6}$ & $2.04 \times 10^{-2}$ & $\mathbf{1.68 \times 10^{-6}}$ & $2.50 \times 10^{-1}$ & $6.99 \times 10^{-6}$ \\
    NARDGP & $7.63 \times 10^{-4}$ & $2.88 \times 10^{-5}$ & $3.34 \times 10^{-2}$ & $1.14 \times 10^{-4}$ & $4.84 \times 10^{-2}$ & $7.16 \times 10^{-2}$ \\
    DGPDF & $1.09 \times 10^{-1}$ & $2.41 \times 10^{-2}$ & $8.81 \times 10^{-4}$ & $1.01 \times 10^{-4}$ & $17.0 \times 10^{-1}$ & $8.13 \times 10^{-1}$ \\
    DGPDFC & $9.99 \times 10^{-4}$ & $2.52 \times 10^{-5}$ & $1.00 \times 10^{-1}$ & $1.37 \times 10^{-4}$ & $\mathbf{3.53 \times 10^{-2}}$ & $2.52 \times 10^{-4}$ \\
    \bottomrule
  \end{tabular}
  \end{adjustbox}
  \label{tab:mse_academic}
\end{table}

We can observe from Table~\ref{tab:mse_academic} that across all scenarios, the multi-fidelity methods outperform the single-fidelity GP by a considerable margin. This is because the multi-fidelity methods take advantage of the additional information the low-fidelity models provide. This shows the advantage of using multi-fidelity GP over single-fidelity GP. One can also conclude from Table~\ref{tab:mse_academic} that methods involving the DGP have higher MSE as compared to the corresponding non-linear autoregressive methods with full GP, specifically when the number of high-fidelity function evaluations is increased. This is because DGP involves the approximation of the posterior distribution using some finite induction points.

We observe from Table~\ref{tab:mse_academic} that the AR1 surrogate demonstrates the lowest MSE compared to the other methods in a linear transformation problem. The high-fidelity function of the linear scaling problem is written as constant scaling of the low-fidelity function with a non-linear additive term. This aligns with the assumed formulation of AR1 as described in Equation \ref{eq:AR1-formulation}, giving AR1 an edge over the other methods. Every other non-linear method except DGPDF also has a low MSE compared to the single-fidelity GP. The non-linear transformation methods involve expanding the surrogate dimensionality by introducing additional dimensions for low-fidelity function evaluations and/or delay terms. The assumed non-linear transformation of the low-fidelity term and the additional dimension of the delay term do not contribute meaningful information to the surrogate model in the case of the linear scaling problem. The additional kernel hyperparameters and the increase in dimension might require more evaluation points to represent the function. Therefore, non-linear auto-regressive methods underperform when compared to the AR1 method. 

Table~\ref{tab:mse_academic} shows that the AR1 surrogate cannot accurately capture non-linear transformations as expected because of the underlying linear formulation. In contrast, other methods tailored for non-linear transformations exhibit effective predictions with low MSE.

In the next example, we discuss the case where the high-fidelity function ($f_h$) and the low-fidelity function ($f_l$) are oscillating functions with phase differences. We observe such cases in our brain where neurons oscillate with certain frequencies where one uses the Hodgekin-Huxley model of them as a surrogate for real data~\cite{hodgkin1952quantitative}.

The high-fidelity function can be expressed as a combination of sine and cosine terms, with the cosine term representing the derivative of the low-fidelity function. Consequently, methods incorporating delay terms are anticipated to outperform others.
\begin{eqnarray*}
f_h(x) &=& \sin{\left(8 \pi x\right)} \cos{\left( \frac{\pi}{10}\right)} + \cos{\left(8 \pi x\right)} \sin{\left( \frac{\pi}{10}\right)} + \cos{\left( 4 \pi x \right)}   \\
&=& \cos{\left( \frac{\pi}{10}\right)} f_l(x) + \sin{\left( \frac{\pi}{10}\right)} \frac{d f_l}{dx} + \cos{\left( 4 \pi x \right)}.
\end{eqnarray*}
NARGP and NARDGP exhibited less accurate fitting to the target high-fidelity curve, as their formulations lack derivatives directly applicable to this example. Conversely, methods incorporating delay terms in surrogate modeling demonstrated accurate predictions by mimicking the derivative of the low-fidelity function. Surprisingly, AR1 also predicts accurately because the phase-shifted oscillation gets simplified into a linear scaling problem which is suitable for AR1. 

DGPDF specifically underperforms for linear-scaling problems and phase-shifted oscillations even when other non-linear autoregressive methods had low MSE. Only one kernel is responsible for capturing all the features of the target high-fidelity function. In other non-linear autoregressive methods, the kernel has a well-defined structure. This underscores the significance of a well-defined kernel structure specifically for DGP.

The three presented cases with different transformations underscore the variability in the performance of different methods across diverse scenarios. This observation emphasizes the necessity of a judicious selection of methods based on the specific characteristics of the modeled system. Prior knowledge about the system can significantly contribute to choosing the most suitable methodology.

\begin{figure}[!t]
    \centering
    \begin{subfigure}{.48\textwidth}
        \centering
        \includegraphics[width=\linewidth, height=0.25\textheight]{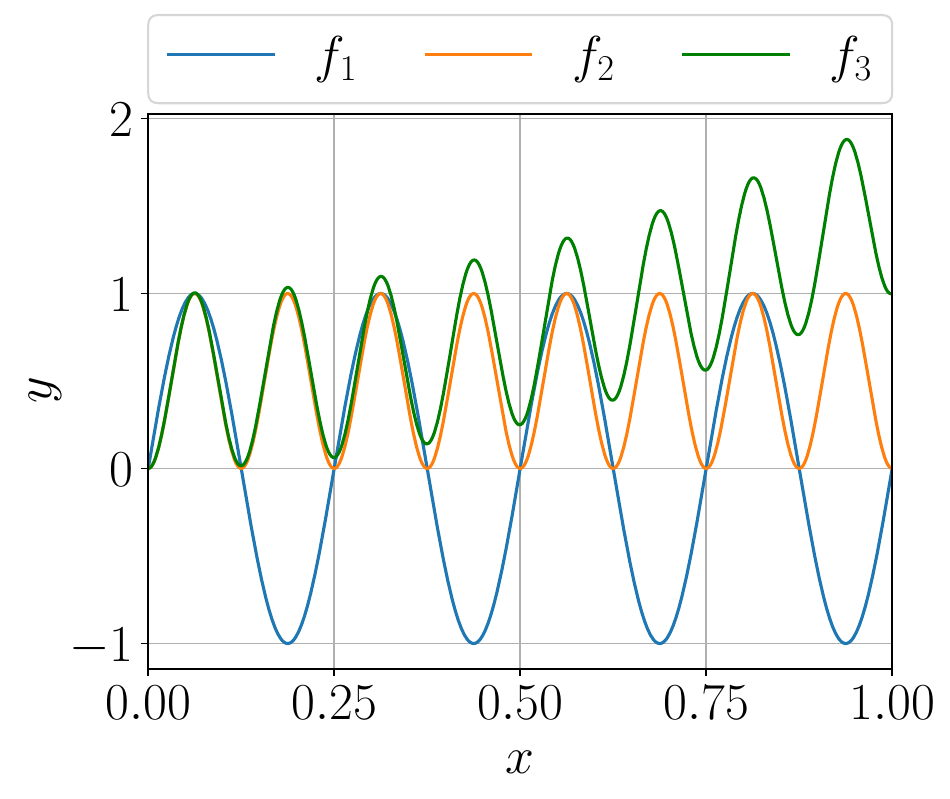}
        \caption{Target functions.}
        \label{fig:function_3fidelities}
    \end{subfigure}
    \begin{subfigure}{.48\textwidth}
        \centering
        \includegraphics[width=\linewidth, height=0.3\textheight]{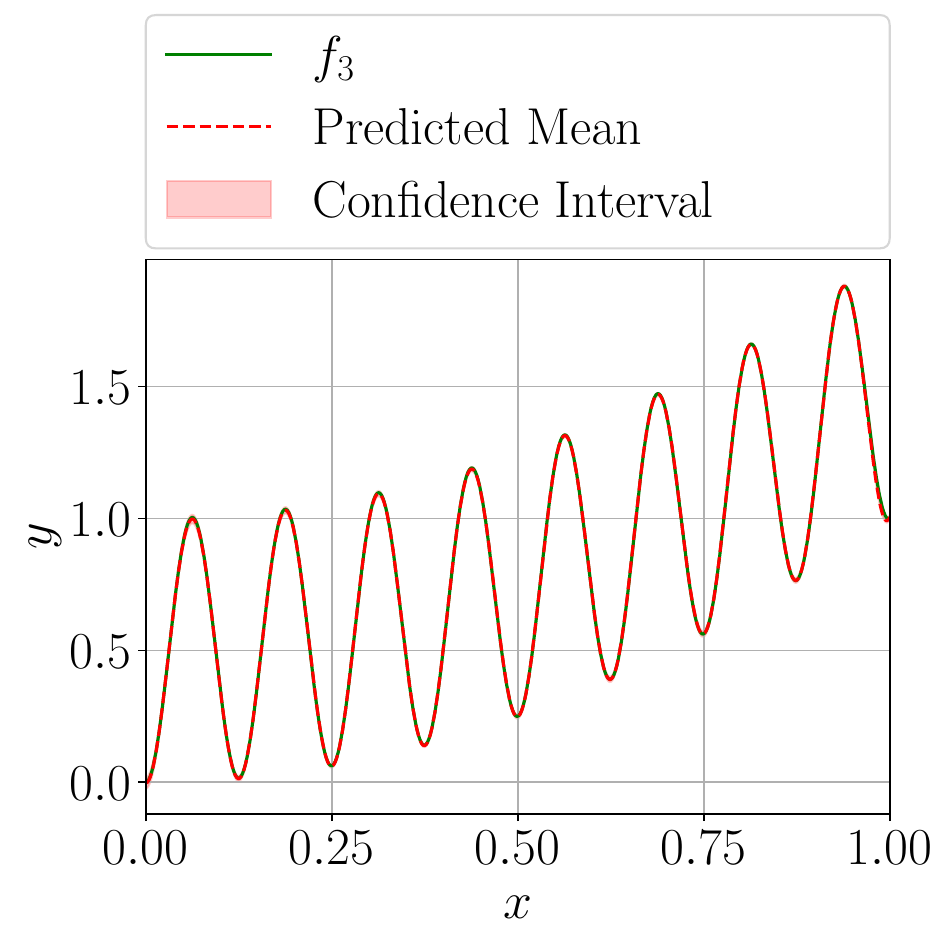}
        \caption{Model prediction for $f_3$ using NARGP.}
        \label{fig:prediction_3fidelities}
    \end{subfigure}
    \caption{An example of multi-fidelity gaussian process surrogates for three fidelity case. We use NARGP to create surrogate because the transformation from first to the second fidelity is non-linear.}
    \label{fig:3fidelities}
\end{figure}
As indicated in Section~\ref{sec:method}, we can extend the surrogate modeling to encompass more than two fidelities. As an example, we consider a three-fidelity case defined as
\begin{eqnarray}
f_1(x) = \sin \left(8 \pi x \right) \nonumber \\ 
f_2(x) = \sin^2 \left(8 \pi x \right) \\ 
f_3(x) = \sin^2 \left(8 \pi x \right) + x^2 \nonumber
\label{eq:3fidelity}
\end{eqnarray}
Figure~\ref{fig:function_3fidelities} shows the target function for the three fidelity cases as described in Equation~\ref{eq:3fidelity}. The transition from the first fidelity to the second fidelity is non-linear, necessitating the use of a non-linear method to construct a surrogate. In this example, we present the results achieved using NARGP. We begin with forty, twelve, and eight randomly sampled points for each fidelity level and then execute the adaptivity algorithm for five steps. We have already observed from the results of two-fidelity cases that the posterior variance for the second fidelity under the same scenario using the NARGP model is very small. Consequently, we can utilize the posterior mean of the surrogate of the second level as the input for the third layer. Figure~\ref{fig:prediction_3fidelities} shows that the surrogate aligns closely with the target function. Analogous to the two-fidelity cases, familiarity with the physical model will guide our selection of the appropriate method.

%% file: sections/results/terra_mech.tex
Terramechanics explores the interaction of wheels with the underlying soil~\cite{bekker1969introduction}. 
There are no exact physical formulas that describe the process of the wheel-soil interaction because of the non-linearity of the pressure-sinkage relations in the soft soil~\cite{dlr189262}. 
Therefore, to simulate wheel movement on soft sands and to estimate forces and torques acting on that wheel, we exploit a wide range of numerical simulations, each one with a different level of discretization, assumptions, accuracy, and computational time. 
This makes terramechanics a good example of multi-fidelity modeling because different numerical models are dedicated to simulating the same physical process, but with different levels of fidelity.

\begin{figure}[!]
    \begin{subfigure}[h]{0.54\linewidth}
        \includegraphics[width=\linewidth]{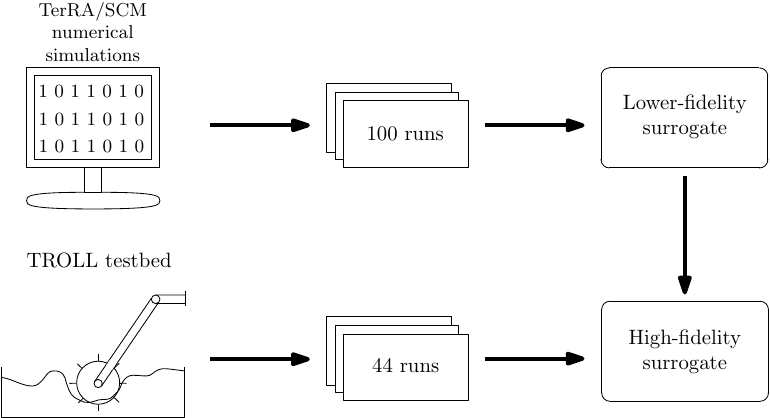} 
     \caption{Setup of the terramechanical experiment. TROLL testbed was used to create 44 high-fidelity runs. Numerical simulations were used to create 100 lower-fidelity runs and train a lower-fidelity surrogate model.}
     \label{fig:terramech_scheme}
    \end{subfigure}
    \hfill
    \begin{subfigure}[h]{0.44\linewidth}
         \includegraphics[width=\linewidth]{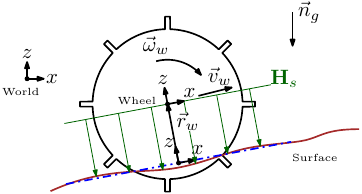}
        \caption{Representation of the input data used in the ML model. $\vec{r}_w$ is a wheel position w.r.t the surface; $\vec{v}_w$ is wheel velocity, $\vec{\omega}_w$ is the angular velocity of the wheel, $\vec{n}_g$ stands for gravity direction, and $\vec{f}_w$ depicts traction force acting on the wheel, our target value we want to predict \cite{fediukov2022multi}.}
        \label{fig:data_sources}
    \end{subfigure}%
    \caption{Different data generators were simulating the wheel-soil interaction for the same scenarios, thus creating an ideal case for multi-fidelity modeling. }
\end{figure}

\begin{table}[!]
  \centering
  \begin{tabular}{ccc}
    \toprule
    \textbf{Variable} & \textbf{Specification} & \textbf{Description} \\
    \midrule
     $\vec{r}_w$           & 3D Input              & Wheel position w.r.t the surface   \\
     $\vec{v}_w$           & 3D Input              & Translational velocity of the wheel      \\
     $\vec{\omega}_w$      & 3D Input              & Angular velocity of the wheel        \\
     $\vec{n}_g$           & 3D Input              & Gravity acting on the wheel   \\
     $f$           & 1D Output             & Traction force acting on the wheel        \\
     \bottomrule
  \end{tabular}
  \caption{Description of the variables from the terramechanical dataset.}
  \label{tab:table_variable}
\end{table}

As data sources for lower-fidelity models for this experiment, we used two types of models, both of which were developed at the Terramechanical lab at German Aerospace Center (DLR): TerRA~\cite{dlr121815} and SCM~\cite{dlr189262}. 
As a high-fidelity data source, DLR's testbed TROLL was used~\cite{dlr121796}.
This testbed was initially created to test and validate simulations for the wheel-soil interactions for extraterrestrial rover projects like MMX~\cite{buse2022mmx} and Scout~\cite{pignede2022toolchain}.
The 44 high-fidelity runs from TROLL were also replicated in the simulation environment to achieve both higher- and lower-fidelity observations on the same set of inputs\footnote{The data is not publicly available but is available upon request.}.
These runs included a variety of different steering and titling angles, different rotation and horizontal velocities, and different movement trajectories and accelerations. 
Movement scenarios for dataset creation were described and justified in the previous paper~\cite{fediukov2022multi}.
The scheme of the overall setup of experiments is depicted in Figure~\ref{fig:terramech_scheme}.

Lower fidelity simulations are iterative simulations with each next step dependent on the state of the previous step, and the output of the simulation depends on several technical variables, which do not have a clear physical meaning and were not included as the input features in the dataset.
The Gaussian process surrogate takes only a subset of the input variables of the simulation into account.
Surrogate GPs were trained to approximate the outputs of TerRA and SCM simulations, and 100 runs were used in both cases.
Each run was 20 seconds long and with a sampling rate of 3 points per second. Runs were conducted with randomly selected surfaces, random sequences of acceleration and deceleration and random sequences of steering commands.
The randomness of a surface is introduced by varying its profile. 
The profile for each run has 6 breaking points, where after each point the change is introduced. Changes are normally distributed with a mean of 0.1 m and a variance of 0.6 m and are gradually introduced at each breakpoint. 
Velocities of the wheel simulation are also distributed normally, with a mean of 1 m/s and variance of 2 m/s.
Steering angle is distributed normally, with 0 mean and 0.5 radians variance.
Both velocities and steering angle were changing each second of the simulation.

We split high-fidelity runs into 29 runs used for training purposes and 15 reserved for the test dataset.
We varied sampling frequency from the high-fidelity data, to verify how performance changes with changes in the amount of high-fidelity data.
Harvesting high-fidelity data is challenging and we want to minimize the damage to the model's performance, therefore we want to analyze the degradation of the prediction accuracy with the decrease in the number of training points.
Each next subset is nested into the subset with a bigger number of training points.
This logic assures the consistency of training points in all down-sampling examples.

All signals were smoothed with a low-pass filter, as in reality spikes of traction force are considered to be noise because only constant application of a force can change the wheel's movement.
The data consists of 12-dimensional input and the objective function is a traction force acting on the wheel.
A description of the variables is detailed in Table~\ref{tab:table_variable}.
A visualization of the wheel and features describing it is depicted in Figure~\ref{fig:data_sources}.

We tested all MF models discussed earlier on the terramechanical data.
MF models shown here were trained with both SCM and TerRA surrogates being the lower-fidelity function and compared with each other. Table~\ref{tab:table_mufintroll} shows the performance of different MF methods w.r.t. the number of high-fidelity data points used during the modeling and numerical simulation which generated the data for the lower-fidelity level.
The performance of the MF models with the TerRA as a lower-fidelity data source shows marginally worse results than with SCM, however still very compatible. 
This is a positive result, given that the TerRA model is much simpler in implementation and less expensive in exploitation.
Changing the number of high-fidelity points can give a hint of the lower acceptable bar when we construct the MF model.

\begin{table}[!]
\caption{Normalized MSE of each MF method for wheel-soil locomotion simulation, w.r.t. number of high-fidelity data points used in training. Simulations using SCM were used as a lower-fidelity layer. The normalized MSE of conventional numerical simulation models is presented for comparison. GP trained only on the high-fidelity points shows good results when we have a lot of data points, but its performance deteriorates with a decrease in the training points. While the MF method, especially NARDGP, can show good results with fewer high-fidelity points and has more consistent performance.}
\begin{adjustbox}{width=\columnwidth, center}
  \begin{tabular}{*{11}{c}}
    \toprule
    & \multicolumn{10}{c}{Number of HF points} \\
    \cmidrule(lr){2-11}
    Method Name & \multicolumn{2}{c}{2085} & \multicolumn{2}{c}{1042} & \multicolumn{2}{c}{521} & \multicolumn{2}{c}{208} & \multicolumn{2}{c}{104}\\
    \cmidrule(lr){2-11}
    & \multicolumn{10}{c}{Numerical simulation used as a lower fidelity layer} \\
     & TerRA & SCM & TerRA 
    & SCM & TerRA & SCM & TerRA & SCM & TerRA & SCM \\
    \midrule
    Single fidelity GP & \multicolumn{2}{c}{0.149} & \multicolumn{2}{c}{0.239} & \multicolumn{2}{c}{0.331} & \multicolumn{2}{c}{0.32} & \multicolumn{2}{c}{0.756} \\
    AR1 & 0.235 & 0.235 & 0.235 & 0.235 & 0.235 & 0.235 & 0.235 & 0.235 & 0.235 & 0.235 \\
    NARGP & 0.184 & 0.211 & 0.2 & 0.23 & 0.218 & 0.726 & 0.251 & 0.256 & 0.724 & 0.726\\ 
    GPDF & 0.41 & 0.207 & 0.327 & 0.231 & 0.3 & 0.218 & 0.25 & 0.233 & 0.724 & 0.724 \\
    GPDFC & 0.41 & 0.207 & 0.327 & 0.231 & 0.3 & 0.218 & 0.25 & 0.233 & 0.724 & 0.724 \\
    NARDGP & $\mathbf{0.116}$ & $\mathbf{0.122}$ & $\mathbf{0.12}$ &  $\mathbf{0.118}$ & $\mathbf{0.123}$ & $\mathbf{0.12}$ & $\mathbf{0.118}$ & $\mathbf{0.117}$ & $\mathbf{0.121}$ & $\mathbf{0.125}$\\
    DGPDF & 0.726 & 0.726 & 0.726 & 0.726 & 0.726 & 0.726 & 0.726 & 0.726 & 0.726 & 0.726\\
    DGPDFC & 0.217 & 0.22 & 0.249 & 0.233 & 0.224 & 0.223 & 0.226 & 0.23 & 0.22 & 0.235\\
    \midrule
    Conventional numerical simulations & \multicolumn{10}{c}{ } \\
    TerRA & \multicolumn{10}{c}{799.92} \\
    SCM & \multicolumn{10}{c}{0.209} \\
    \bottomrule
  \end{tabular}
  \label{tab:table_mufintroll}
 \end{adjustbox}
\end{table} 

\begin{figure}
  \centering
      \begin{center}
        \includegraphics[width=0.7\textwidth]{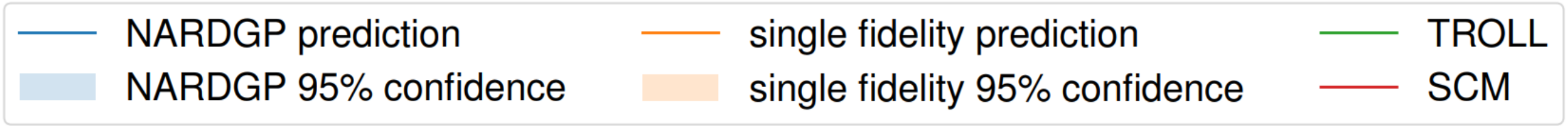}
    \end{center}
      \begin{subfigure}{\textwidth}
        \centering
        \includegraphics[width=\linewidth, height=0.2\textheight]{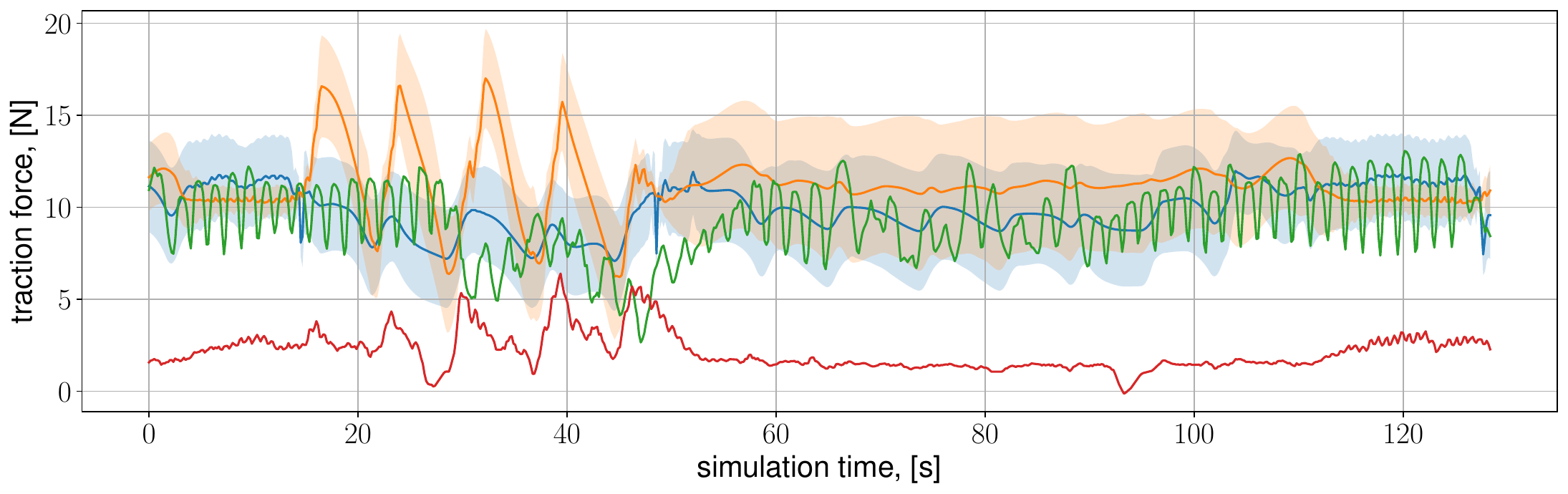}
        \label{fig:nardgp_comparison1}
    \end{subfigure} 
    \newline
      \begin{subfigure}{\textwidth}
        \centering
        \includegraphics[width=\linewidth, height=0.2\textheight]{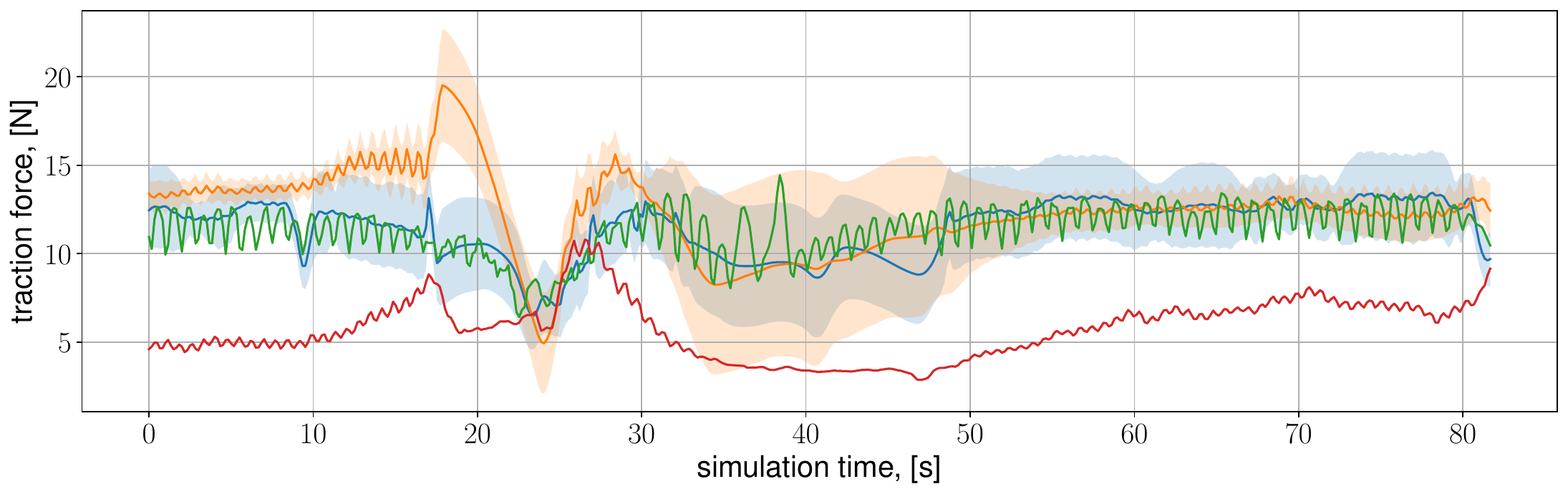}
        \label{fig:nardgp_comparison2}
    \end{subfigure} 
    \newline
      \begin{subfigure}{\textwidth}
        \centering
        \includegraphics[width=\linewidth, height=0.2\textheight]{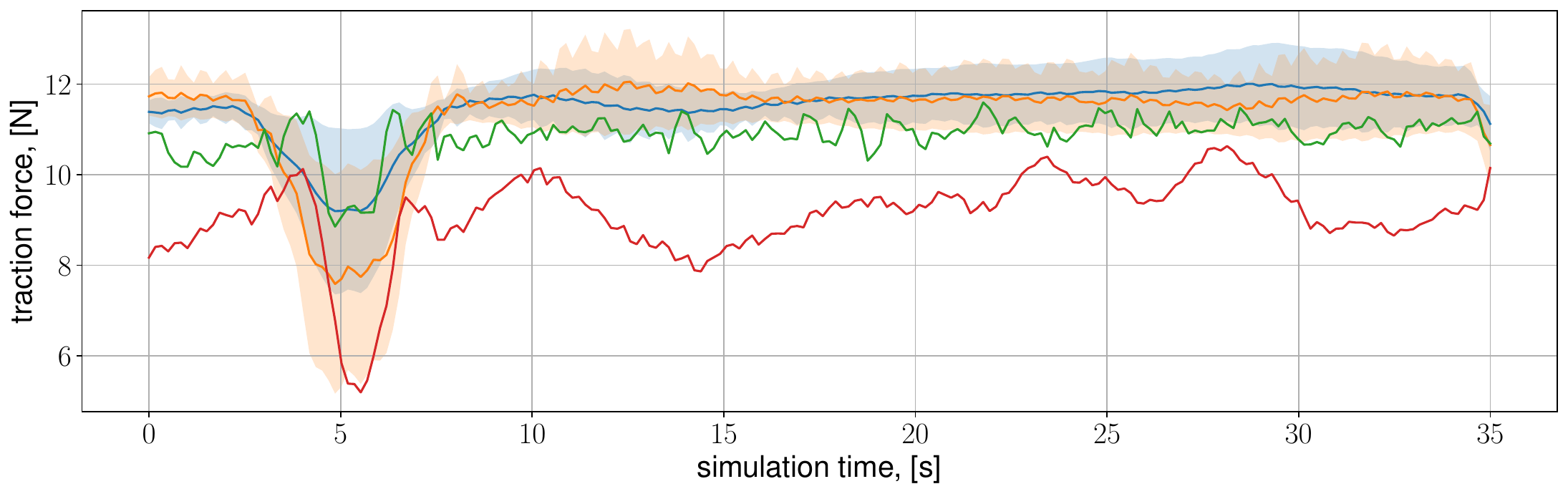}
        \label{fig:nardgp_comparison3}
    \end{subfigure} 
     \caption{Visual comparison of the best performing MF ML model, NARDGP, on several test runs. 
    It is compared against the ground truth from the TROLL sensors, single-fidelity model and the most popular numerical simulation SCM. 
    The overall performance is reasonably better compared to other methods, but the model is visibly over-confident in its uncertainty estimation.}
\label{fig:nardgp_comparison}
\end{figure}

Models comparison with different subsets is illustrated in Table~\ref{tab:table_mufintroll}.
As we can see there, the multi-fidelity approach can handle the drastic decrease in the number of used high-fidelity points, while the single-fidelity model's performance decreases, as expected.
Several examples of best-performing methods, NARDGP, are shown in Figure~\ref{fig:nardgp_comparison}.

\begin{figure}
  \centering
      \begin{subfigure}{\textwidth}
        \centering
        \includegraphics[width=\linewidth, height=0.2\textheight]{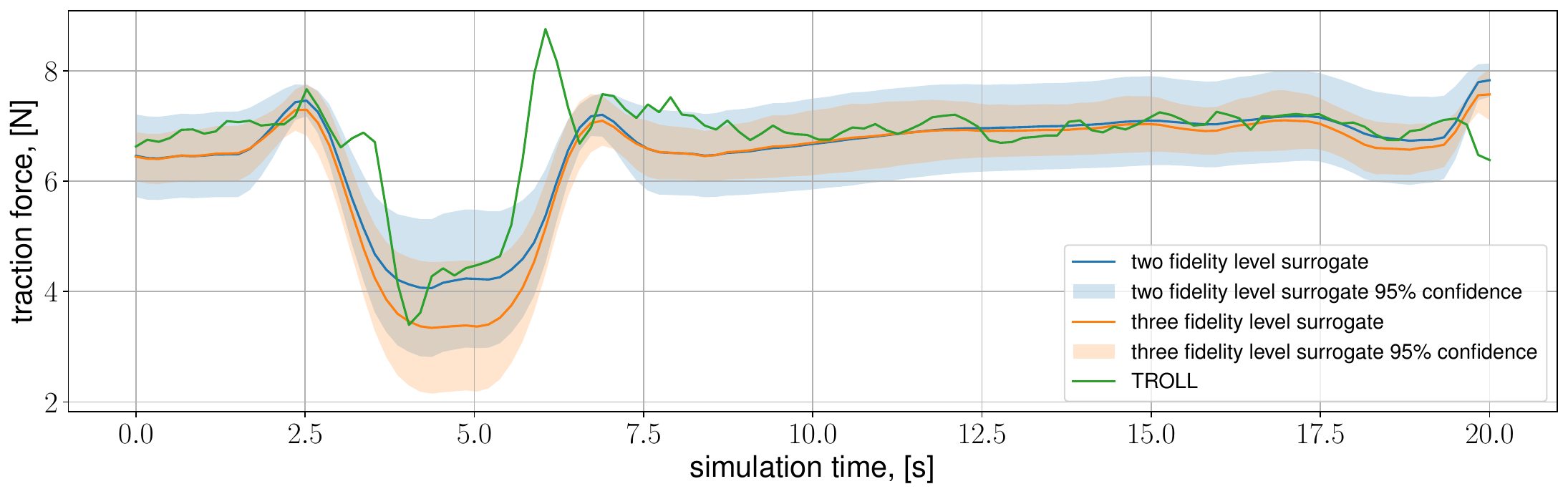}
        \label{fig:nardgp_comparison1}
    \end{subfigure} 
    \newline
      \begin{subfigure}{\textwidth}
        \centering
        \includegraphics[width=\linewidth, height=0.2\textheight]{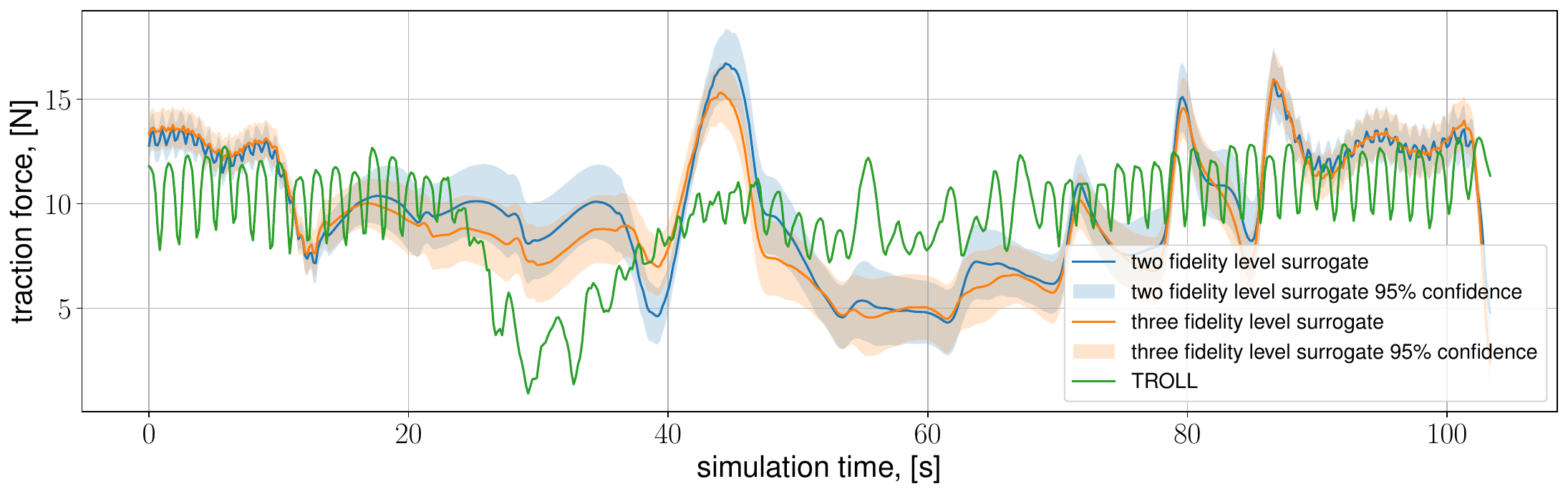}
        \label{fig:nardgp_comparison2}
    \end{subfigure} 
     \caption{Visual comparison of the two- and three-level NARDGP surrogate performance against the test runs from the high-fidelity TROLL experiments. The results are not identical, but the performance increase is not promising, therefore we put aside proper testing of three-level MF ML surrogates for terramechanical data.}
\label{fig:two_vs_three_comparison}
\end{figure}

Empirically, there is no advantage in building more than two layers of fidelity for this regression task, as SCM simulation covers all the abilities of TerRA simulation and gives much more reliable results~\cite{dlr189262}.
We conducted a comparison of the two-level NARDGP model performance where SCM data serves as a lower-fidelity and TROLL as the high-fidelity data source and three-level NARDGP, with new lower-fidelity level from TerRA data, so that SCM simulations become medium-level.
For both cases, we used 43 high-fidelity data points from TROLL experiments, as we are first of all interested in the model's performance with as few high-fidelity points as possible, and 1500 data points from SCM simulation. 
For the three-level model, we sampled 3000 data points from the TerRA simulation.
Two of the examples are depicted in Figure~\ref{fig:two_vs_three_comparison} and they show a very limited difference between the performance of the two approaches. 
Given the limited scale of improvements we didn't conduct a full analysis as in Table~\ref{tab:table_mufintroll}, but we will leave this for further work.

\begin{figure}
  \centering
    \includegraphics[scale=0.6]{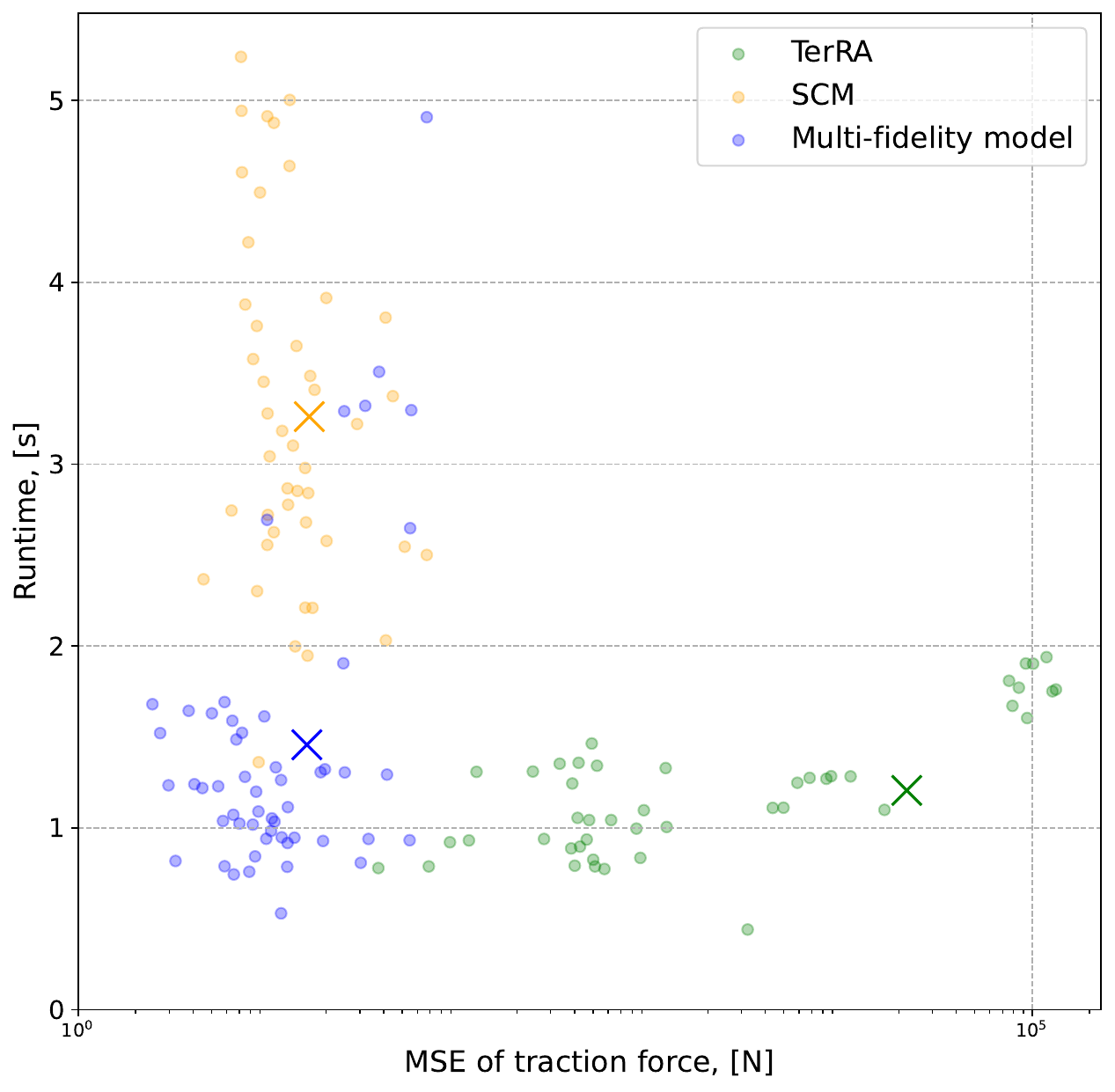} 
     \caption{Mean squared error comparison of predictions from the conventional numerical simulation methods and MF ML model. The multi-fidelity approach is computationally efficient while maintaining a high level of precision.}
\label{fig:comparison_plot}
\end{figure}

One of the main advantages of MF for complex systems modeling is that it can decrease computational and time costs without decreasing prediction accuracy.
MF ML model outperforms both conventional numerical simulations TerRA and SCM simultaneously by MSE and by the runtime.
Figure~\ref{fig:comparison_plot} shows 44 runs for both the MF ML model and two baseline numerical simulations (TerRA and SCM), distributed by their MSE and runtime. 
As we can see, the performance of the MF ML model is indeed faster and more accurate than the conventional terramechanical models.

\begin{table}[htbp]
\caption{Mean pinball loss measures the errors of over- and under-confidence for each quantile individually~\cite{fasiolo2021fast}. Negative loglikelihood (NLL) measures how likely observed data was generated by the model~\cite{song2019distribution}. Expected normalized calibration error (ENCE) orders and splits instances by variance into bins and calculates the scaled difference between the predicted error and variance in each bin~\cite{levi2022evaluating}. Mean predictive interval width (MPIW) measures the sharpness of the prediction and is defined for each confidence interval $\alpha$ as a width of this interval~\cite{yao2019quality}.
Normal and isotonic calibration have similar results almost by every metric, but normal calibration has a significantly worse sharpness metric, which leaves us only with Isotonic regression.}
    \begin{adjustbox}{center}
      \begin{tabular}{*{5}{l}}
        \toprule
        & \multicolumn{4}{c}{Calibration metrics} \\
        Calibration method & Pinball & NLL & ENCE & MPIW \\
        \midrule
        Uncalibrated & 0.696 & 6.648 & 2.232 & \textbf{0.093} \\
        Isotonic regression & \textbf{0.424} & \textbf{1.932} & 0.237 & 0.422\\ 
        Normal calibration & \textbf{0.424} & 2.229 & \textbf{0.21} & 175.675\\
        Beta calibration & 0.484 & 2.74 & 0.62 & 0.187\\
        \bottomrule
      \end{tabular}
  \label{tab:table_calibr}
  \end{adjustbox}
\end{table}

\begin{figure}
  \centering
      \begin{subfigure}{\textwidth}
        \centering
        \includegraphics[width=\linewidth]{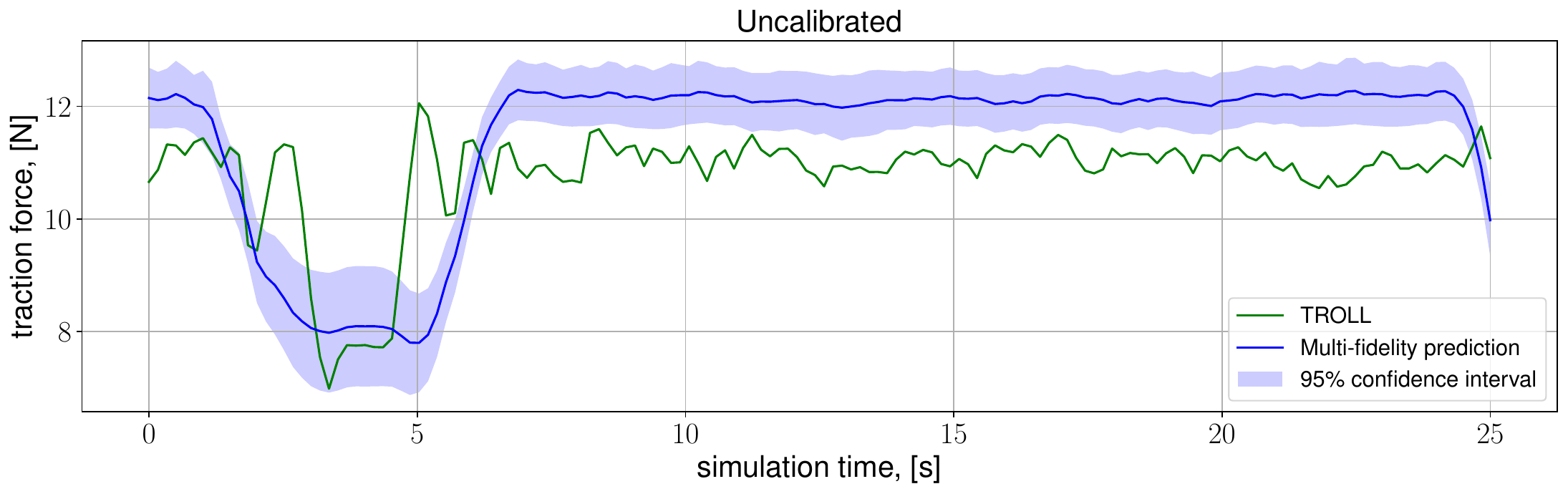}
        \label{fig:calibration1}
    \end{subfigure} 
    \newline
      \begin{subfigure}{\textwidth}
        \centering
        \includegraphics[width=\linewidth]{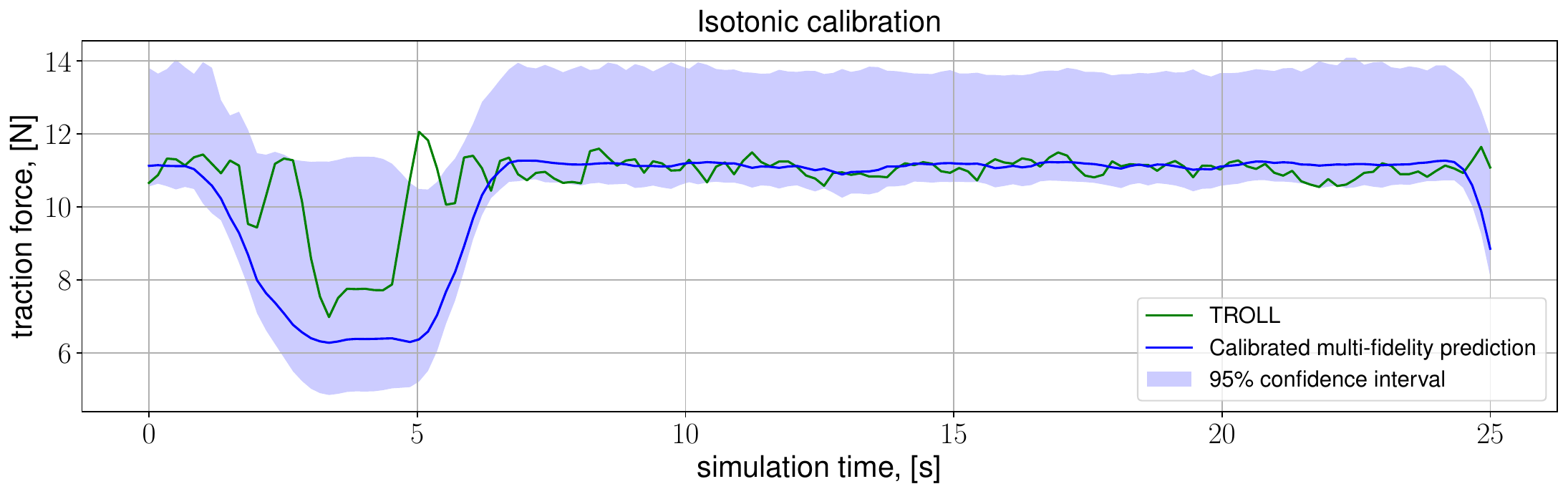}
        \label{fig:calibration2}
    \end{subfigure} 
     \caption{Original uncalibrated prediction is visibly overconfident because the ground truth traction force from TROLL quite often finds itself outside even the 95\% confidence interval. By applying Isotonic regression, we achieve quantile calibrated prediction and here predicted confidence intervals capturing ground truth much better.}
\label{fig:calibration}
\end{figure}

Another advantage of the Gaussian process is the implicit handling of uncertainties. 
Due to its Bayesian nature, we can propagate uncertainties easily using the \RefEq{eqn:posterior_cov}. 
It is an important part of predicting complex simulation systems, both for operational and research purposes.
This adds more depth to the understanding and explainability of predictions and makes ML black boxes more transparent.

As we can see in Figure~\ref{fig:nardgp_comparison}, the main disadvantage of the NARDGP multi-fidelity prediction is that confidence intervals are too narrow, as discussed in Section~\ref{sec:calibration}.
This is a critical flaw when it comes to deploying ML models in practice, especially for highly sensitive operations like extraterrestrial rovers, where the cost of error is very high due to the inability to quickly compensate for the wrong move or repair the rover.
To solve this problem, we applied several calibration methods, mentioned in Section~\ref{sec:calibration}.
Isotonic regression is an algorithm enabling quantile calibration, inspired by a Platt scaling algorithm from classification~\cite{kuleshov2018accurate}.
Beta calibration was chosen as an algorithm for achieving distribution calibration~\cite{song2019distribution}.
Normal calibration~\cite{kuppers2022parametric} introduces a scaling parameter for the predicted variance, preserving a Gaussian nature of the posterior prediction. 
A comparison of all calibration techniques on all of the test runs can be seen in Table~\ref{tab:table_calibr}. 
As an example, we took NARDGP trained with 41 high-fidelity data points and calibrated it with the calibration mentioned above techniques.
From this table, we can see that isotonic regression performs the best calibration, despite that it is only a global calibration of the marginal distribution.
This means that the NARDGP model was constantly making overconfident predictions and global scaling of the variance solves the problem. 
These findings are consistent with the previous study, researching Gaussian process regression calibration for terramechanical data, but conducted only in the context of single-level medium-fidelity SCM data~\cite{huhne2023uncertainty}. 
An example of a calibration on one test run could be seen in Figure~\ref{fig:calibration}, where prediction became less overconfident and 95\% internal captures almost the entire actual observed traction force signal.  

%% file: sections/results/plasma.tex
Harnessing energy from plasma fusion promises to offer a clean alternative energy source. To achieve this goal, creating a self-sustained burning plasma is essential. Achieving a sustained burning plasma remains elusive due to physical and technological challenges. One prominent physical hurdle is the occurrence of small-scale fluctuations in confined plasma, which lead to energy loss. These small-scale fluctuations are referred to as microturbulence. Mitigating energy losses caused by microturbulence is a significant challenge within the plasma physics community.

\begin{figure}[!t]
    \begin{center}
        \includegraphics[width=.8\textwidth]{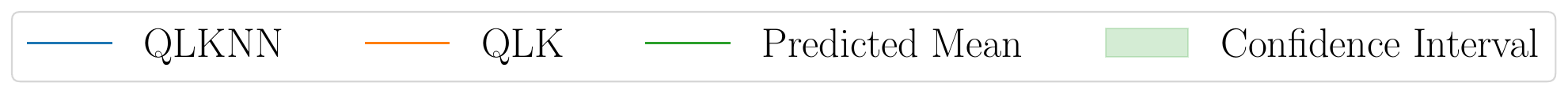}
    \end{center}
    \begin{subfigure}{.46\textwidth}
        \centering
        \includegraphics[width=\linewidth, height=0.18\textheight]{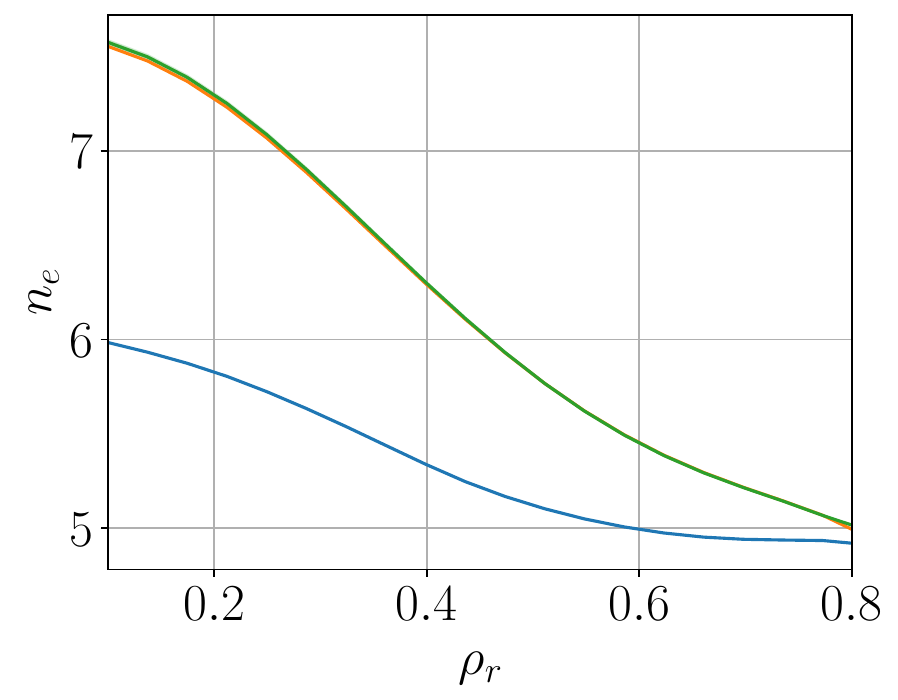}
        \caption{$n_e$}
        \label{fig:ne-nargp}
    \end{subfigure}
    \begin{subfigure}{.46\textwidth}
        \centering
        \includegraphics[width=\linewidth, height=0.18\textheight]{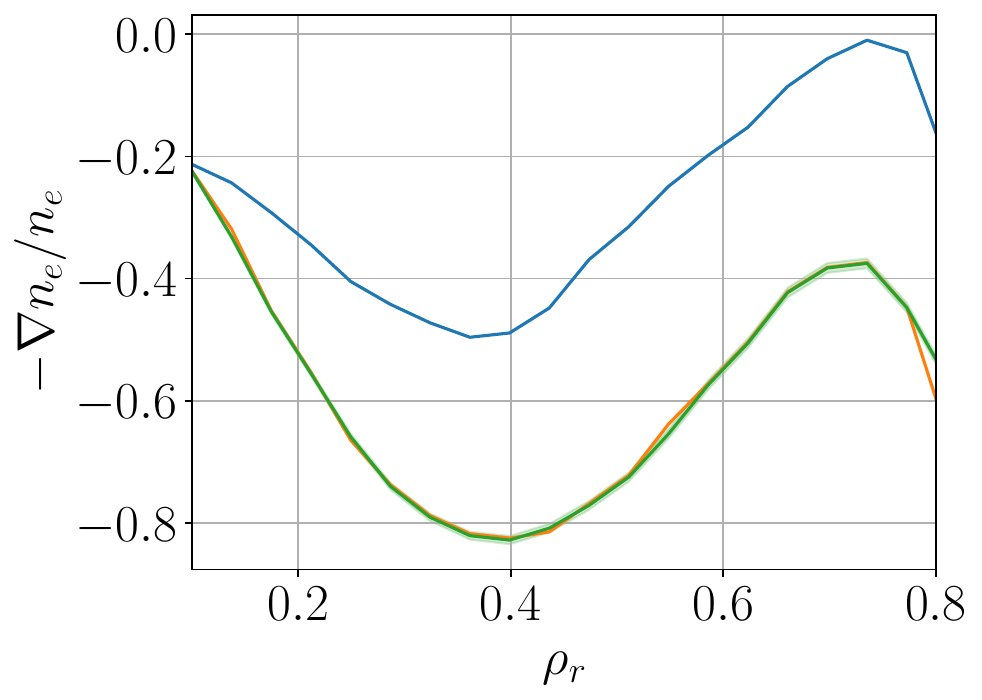}
        \caption{$-\nabla n_e/n_e$}
        \label{fig:log_der_ne-nargp}
    \end{subfigure}
    \newline
    \begin{subfigure}{.46\textwidth}
        \centering
        \includegraphics[width=\linewidth, height=0.18\textheight]{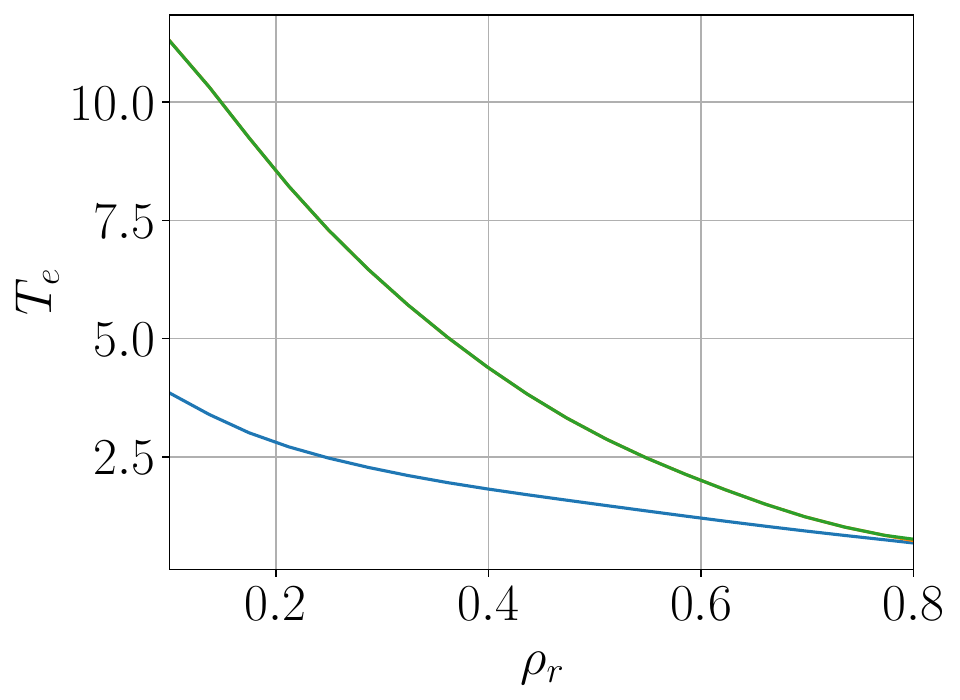}
        \caption{$T_e$}
        \label{fig:Te-nargp}
    \end{subfigure} 
    \begin{subfigure}{.46\textwidth}
        \centering
        \includegraphics[width=\linewidth, height=0.18\textheight]{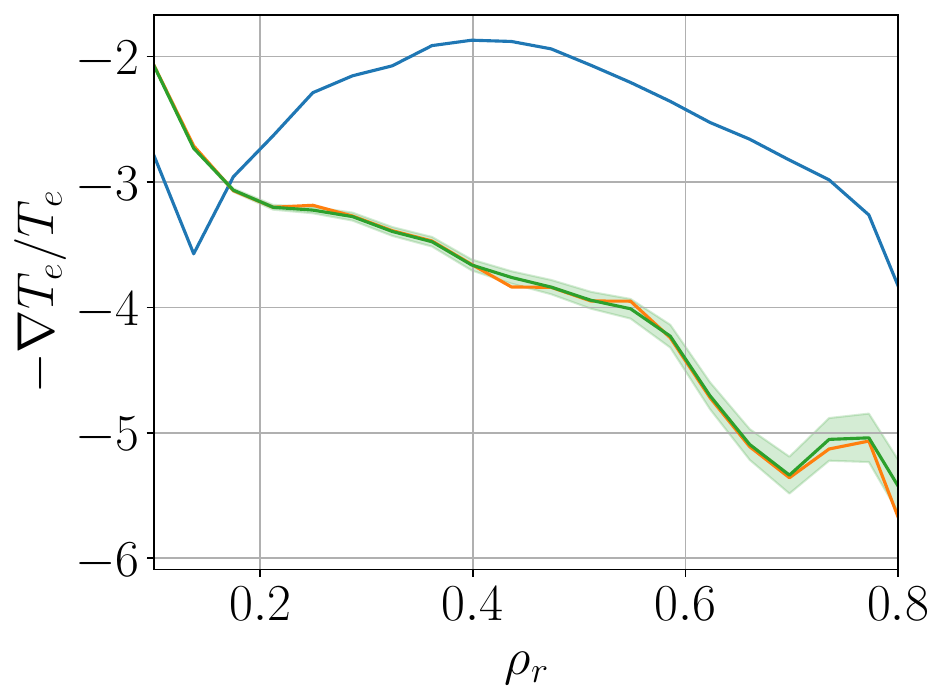}
        \caption{$-\nabla T_e/T_e$}
        \label{fig:log_der_Te-nargp}
    \end{subfigure}
    \newline 
    \begin{subfigure}{.46\textwidth}
        \centering
        \includegraphics[width=\linewidth, height=0.18\textheight]{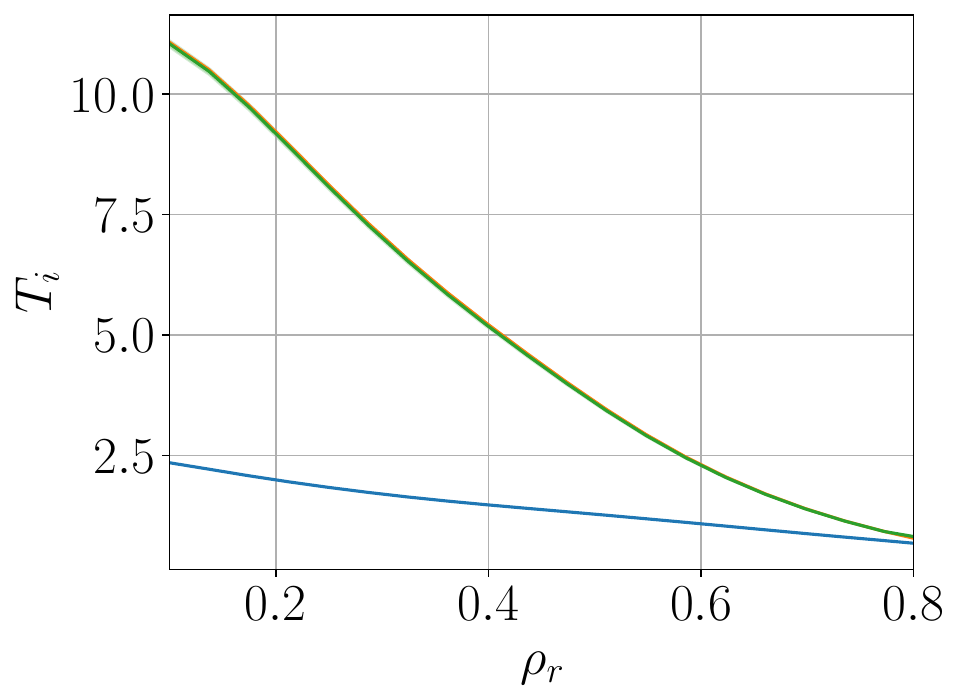}
        \caption{$T_i$}
        \label{fig:Ti-nargp}
    \end{subfigure}  
    \begin{subfigure}{.46\textwidth}
        \centering
        \includegraphics[width=\linewidth, height=0.18\textheight]{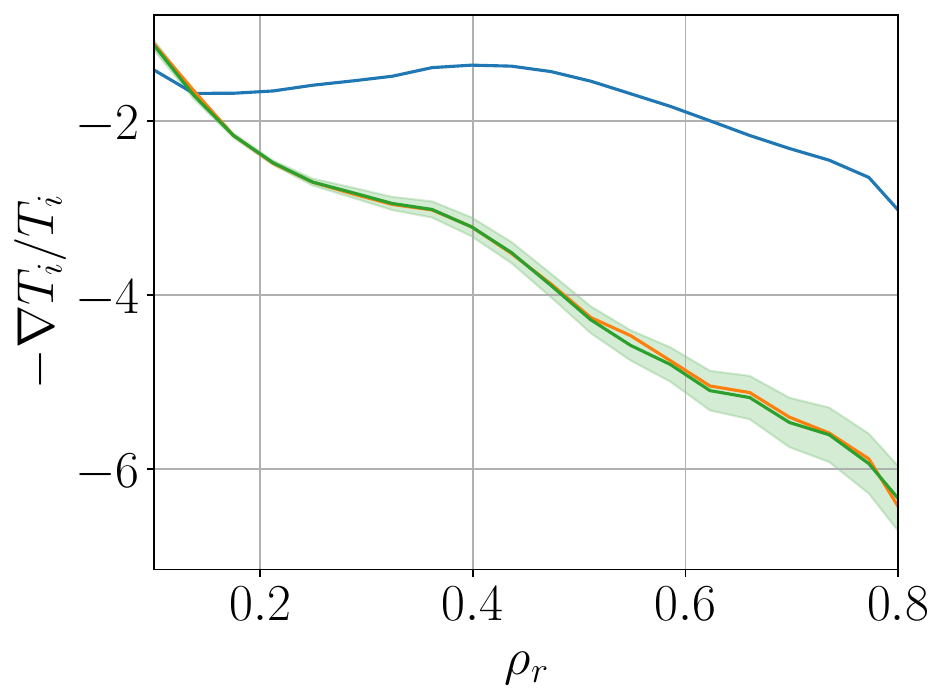}
        \caption{$-\nabla T_i/T_i$}
        \label{fig:log_der_Ti-nargp}
    \end{subfigure}
    \caption{Steady state solution for different quantities simulated for the shot \#34954. Parameters other than radial distance ($\rho_r$) are set as $1$. The blue and orange curves show the simulation results for QLKNN (low-fidelity) and QLK (high-fidelity), respectively. Green curves show the predicted mean and 95\% confidence interval for the NARGP surrogate.}
    \label{fig:plasma-figure}
\end{figure}

This work simulates plasma fusion using the Automated System for TRansport Analysis (ASTRA)~\cite{pereverzev2002astra, fable2013novel} modeling suite. ASTRA solves four 1-dimensional transport equations for electron density ($n_e$), electron temperature ($T_i$), ion temperature ($T_i$) and poloidal flux ($\Psi$). Interested readers can refer to~\cite{hinton1976theory} for details. It uses different turbulent solvers to predict the density and temperature profiles given some initial conditions (initial profiles of $T_i$, $T_e$, and $n_e$) and the last-closed-flux surface. The calculation of the turbulent flux is one of the computationally expensive parts of the code. One can couple ASTRA with different turbulent solvers. In this work, we will use QuaLiKiz (QLK)~\cite{bourdelle2015turbulent} and QuaLiKiz Neural Network (QLKNN)~\cite{ho2021neural, van2020fast} as the two subroutines. QLK is a quasi-linear turbulence solver for the linearized gyrokinetic Vlasov equation. QLKNN is a neural network surrogate trained on a 10-dimensional Latin hypercube for a quick flux evaluation~\cite{van2020fast}. ASTRA simulation with QLK is the high-fidelity model, whereas ASTRA with QLKNN is the low-fidelity model.

We simulate a plasma discharge until it reaches a steady state. At the steady state, the heat and particle fluxes match the integrated sources at all the radial locations. In this work we simulate shot \#34954 
until it reaches a steady state. We train the surrogate on a six-dimensional space mentioned in Table~\ref{tab:param_plasma}. We create six surrogates for six different quantities, namely, steady-state $T_i$, $T_e$, and $n_e$, and the corresponding negative derivative of the logarithm of each quantity ($-\nabla T_i/ T_i$, $-\nabla T_e/ T_e$, and $-\nabla n_e/n_e$). We normalize each quantity before training the surrogate. We use 40 high-fidelity and 400 low-fidelity training points. 

Figure~\ref{fig:plasma-figure} shows the simulation results for QLK~(high-fidelity model) and QLKNN~(low-fidelity model) when parameters other than radial distance ($\rho_r$) are set as $1$. The figure also shows the predicted results for multi-fidelity surrogates using NARGP. Table~\ref{tab:mse_plasma} shows the spatial average of mean square error for each surrogate.

We observe that non-linear auto-regressive multi-fidelity surrogates perform better than the single-fidelity GP. However, the difference between single-fidelity GP and multi-fidelity surrogates is not too significant. This may be due to the underlying simple structure that single-fidelity GP was able to learn relatively easily. AR1 struggles to model the quantity $-\nabla n_e/n_e$, which suggests that the relation between low-fidelity and high-fidelity could be non-linear for the case of $-\nabla n_e/n_e$. Moreover, DGPDF also has a high MSE for $-\nabla n_e/n_e$, whereas DGPDFC fits the quantity accurately. This re-iterates the significance of structured kernels, especially for deep Gaussian processes.

\begin{table}[!t]
    \caption{List of parameters for plasma microturbulence surrogate modeling.}
    \centering
    \begin{tabular}{cc}
        \toprule
        \textbf{Parameter Name} &  \textbf{Range} \\
        \midrule
        Radial distance ($\rho_r$) & $[0.1, 0.8]$ \\ 
        Initial $T_i$ scaling & $[0.9, 1.1]$ \\
        Initial $T_e$ scaling & $[0.9, 1.1]$ \\
        Initial $n_e$ scaling & $[0.9, 1.1]$ \\
        Toroidal velocity scaling & $[0.9, 1.1]$ \\
        Safety factor scaling & $[0.9, 1.1]$\\
        \bottomrule
    \end{tabular}
    \label{tab:param_plasma}
\end{table}

\begin{table}[!]
  \caption{Spatial average value of MSE for different surrogates across different quantities for the plasma microturbulence surrogate modelling.}
  \begin{adjustbox}{width=\columnwidth,center}
  \begin{tabular}{*{7}{c}}
    \toprule
    & \multicolumn{6}{c}{Quantity of Interest} \\
    \cmidrule(lr){2-7}
    Method Name& $T_i$ & $T_e$ & $n_e$ & $-\nabla T_i/T_i$ & $-\nabla T_e/T_e$ & $-\nabla n_e/n_e$ \\
    \midrule
    Single-fidelity GP & $8.57 \times 10^{-3}$ & $8.68 \times 10^{-4}$ & $1.07 \times 10^{-2}$ & $2.26 \times 10^{-1}$ & $1.04 \times 10^{-1}$ & $7.83 \times 10^{-2}$ \\
    AR1 & $1.73 \times 10^{-2}$ & $2.26 \times 10^{-3}$ & $3.31 \times 10^{-2}$ & $4.89 \times 10^{-1}$ & $8.21 \times 10^{-2}$ & $1.33$ \\
    NARGP & $4.46 \times 10^{-3}$ & $\mathbf{8.13 \times 10^{-4}}$ & $7.12 \times 10^{-3}$ & $1.39 \times 10^{-1}$ & $\mathbf{6.9 \times 10^{-2}}$ & $\mathbf{9.57 \times 10^{-3}}$ \\ 
    GPDF & $3.07 \times 10^{-3}$ & $\mathbf{8.13 \times 10^{-4}}$ & $6.3 \times 10^{-3}$ & $1.94 \times 10^{-1}$ & $\mathbf{6.9 \times 10^{-2}}$ & $1.44 \times 10^{-2}$ \\ 
    GPDFC & $3.07 \times 10^{-3}$ & $\mathbf{8.13 \times 10^{-4}}$ & $6.3 \times 10^{-3}$ & $1.94 \times 10^{-1}$ & $\mathbf{6.9 \times 10^{-2}}$ & $1.44 \times 10^{-2}$ \\ 
    NARDGP & $2.89 \times 10^{-3}$ & $8.56 \times 10^{-4}$ & $6.15 \times 10^{-3}$ & $4.41 \times 10^{-1}$ & $1.65 \times 10^{-1}$ & $1.72 \times 10^{-1}$ \\
    DGPDF & $4.52 \times 10^{-3}$ & $8.58 \times 10^{-4}$ & $\mathbf{5.42 \times 10^{-3}}$ & $1.83 \times 10^{-1}$ & $8.06 \times 10^{-2}$ & $3.57$ \\
    DGPDFC & $\mathbf{2.88 \times 10^{-3}}$ & $8.58 \times 10^{-4}$ & $\mathbf{5.42 \times 10^{-3}}$ & $\mathbf{1.38 \times 10^{-1}}$ & $1.82 \times 10^{-1}$ & $4.37 \times 10^{-2}$ \\
    \bottomrule
  \end{tabular}
  \end{adjustbox}
  \label{tab:mse_plasma}
\end{table}

%% file: sections/conclusion.tex
This paper describes different multi-fidelity Gaussian process surrogate modeling methods. We extend non-linear autoregressive models for full GP to accommodate cases that involve more than two fidelities. Using a structured kernel with delay terms, we also suggest a new family of multi-fidelity GP models (GPDFC and DGPDFC). Finally, we test all the modeling methods on different academic and real-world problems. 

We observe that not all the multi-fidelity methods performed well under all the scenarios. The quality of prediction depends upon the underlying relationship between the low and high-fidelity models. Prior knowledge about that would help us choose the correct modeling method. In many scenarios, that relationship is not known. Therefore, after constructing the surrogate, one must carefully validate its predictions. We observe that the structured kernel significantly improves models involving deep Gaussian processes.

We also conclude that the multi-fidelity simulations can preserve the performance of higher-fidelity simulations but reach the speed and cost of low-fidelity ones. This can aid different research fields in analyzing the underlying system or improving the algorithm using outer loop methods~\cite{peherstorfer2018survey}.

%% file: sections/acknowledgement.tex
Research by Vladyslav Fediukov, Michael Bergmann and Kislaya Ravi is funded by the Helmholtz
Association under the ”Munich School for Data Science - MuDS”(HIDSS-006). 
Felix Dietrich acknowledges funding by the DFG project no. 468830823, and also association to DFG-SPP-229.

%% file: sections/appendices.tex
\vspace{-2em}
\begin{figure}[H]
    \begin{center}
        \includegraphics[width=.8\textwidth]{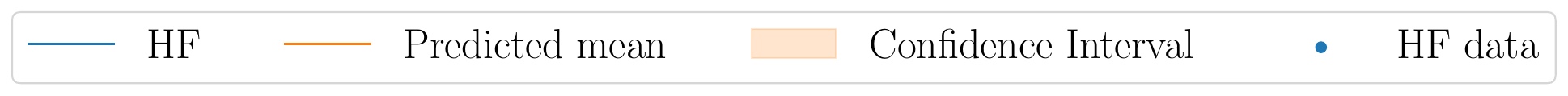}
    \end{center}
    \begin{subfigure}{.46\textwidth}
        \centering
        \includegraphics[width=\linewidth, height=0.18\textheight]{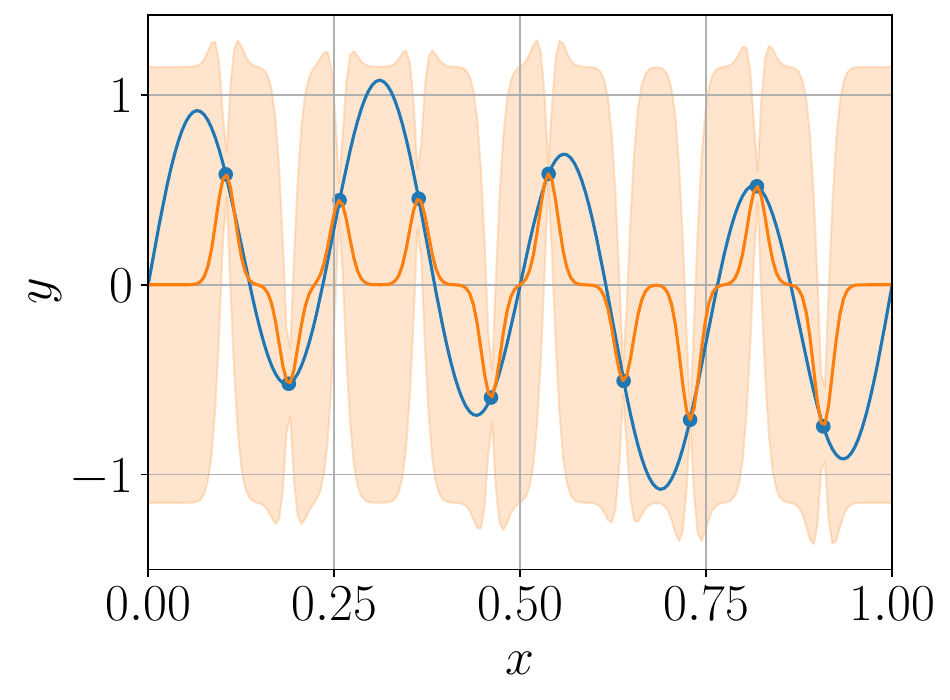}
        \caption{GP}
        \label{fig:linear-gp}
    \end{subfigure} 
    \begin{subfigure}{.46\textwidth}
        \centering
        \includegraphics[width=\linewidth, height=0.18\textheight]{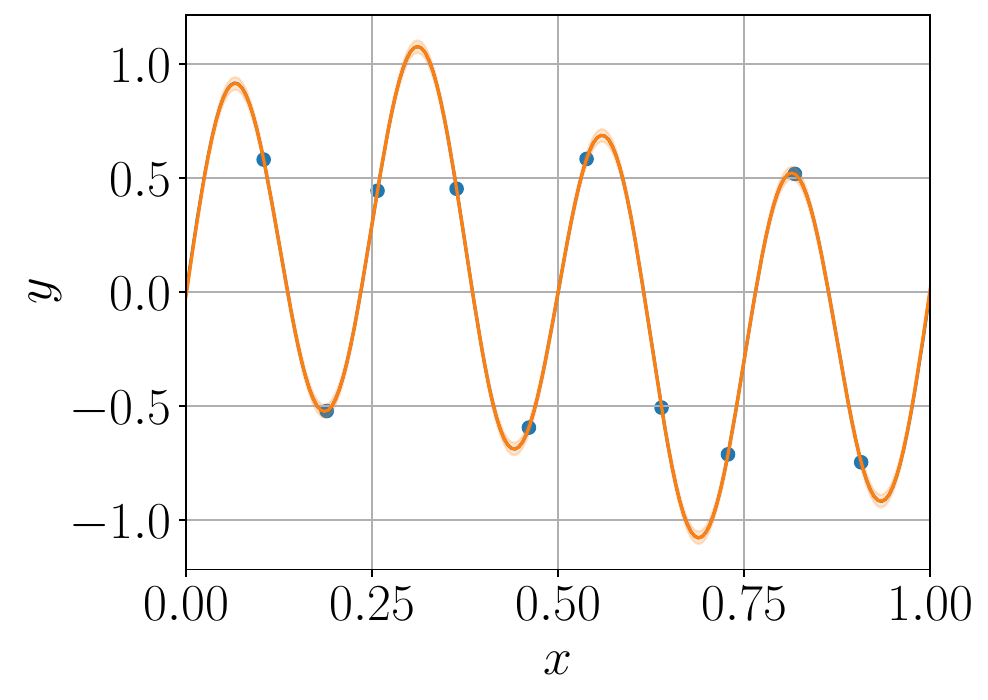}
        \caption{AR1}
        \label{fig:linear-ar1}
    \end{subfigure} 
    \newline 
    \begin{subfigure}{.46\textwidth}
        \centering
        \includegraphics[width=\linewidth, height=0.18\textheight]{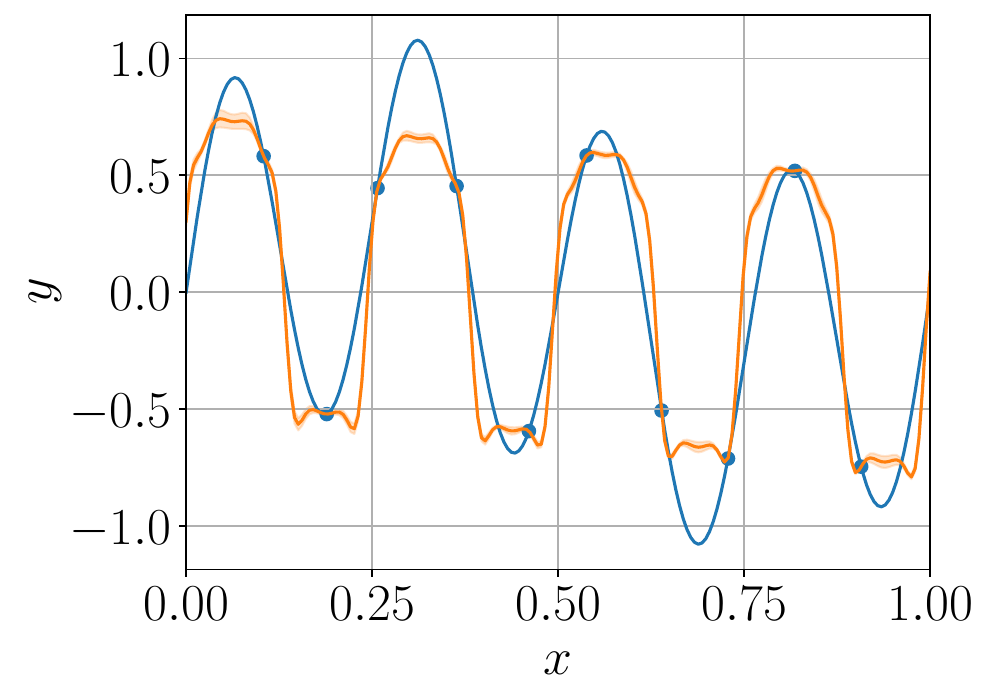}
        \caption{NARGP}
        \label{fig:linear-nargp}
    \end{subfigure}
    \begin{subfigure}{.46\textwidth}
        \centering
        \includegraphics[width=\linewidth, height=0.18\textheight]{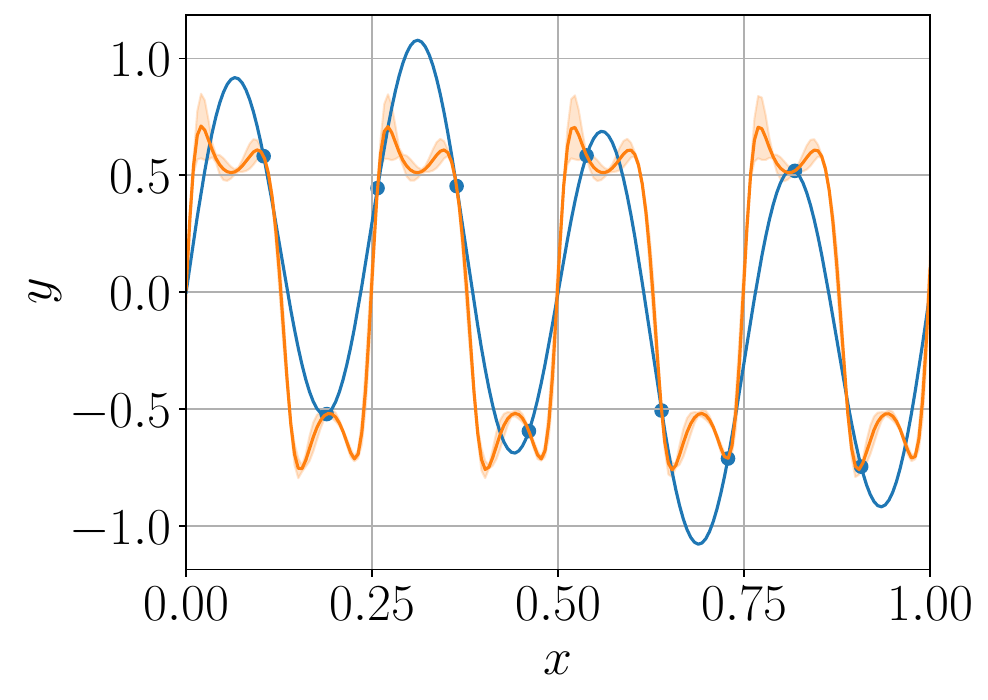}
        \caption{GPDF}
        \label{fig:linear-gpdf}
    \end{subfigure}
    \newline 
    \begin{subfigure}{.46\textwidth}
        \centering
        \includegraphics[width=\linewidth, height=0.18\textheight]{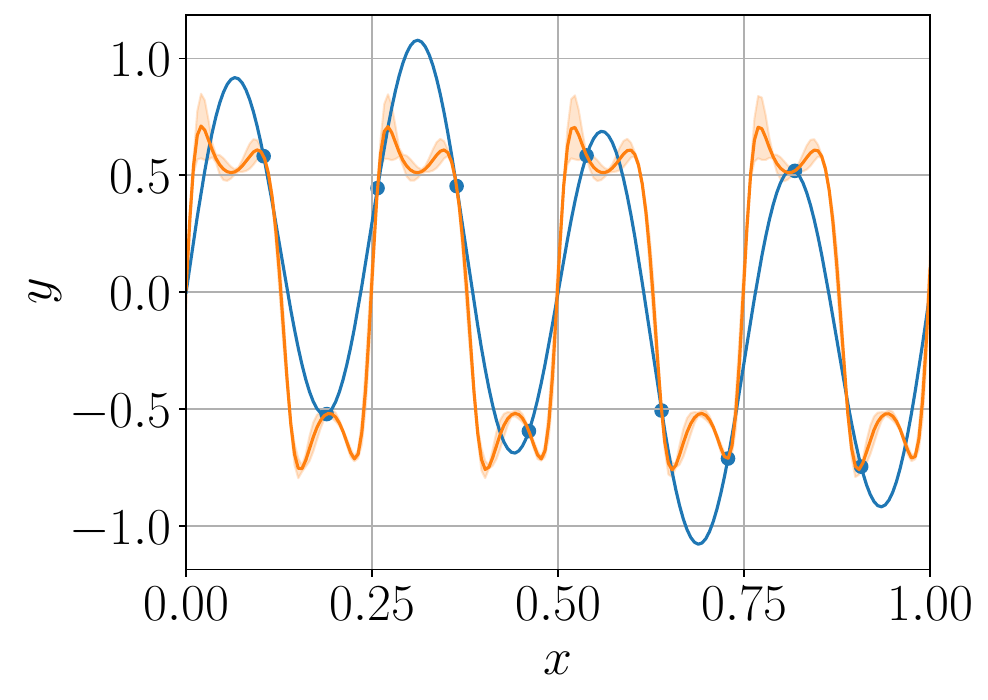}
        \caption{GPDFC}
        \label{fig:linear-gpdfc}
    \end{subfigure}
    \begin{subfigure}{.46\textwidth}
        \centering
        \includegraphics[width=\linewidth, height=0.18\textheight]{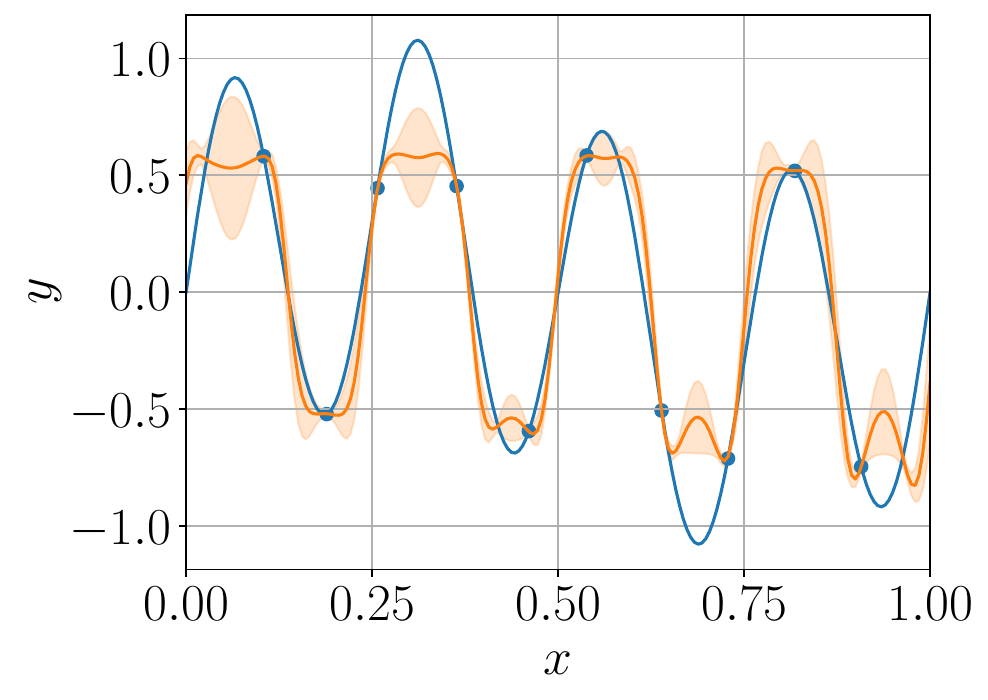}
        \caption{NARDGP}
        \label{fig:linear-nardgp}
    \end{subfigure}
    \newline 
    \begin{subfigure}{.46\textwidth}
        \centering
        \includegraphics[width=\linewidth, height=0.18\textheight]{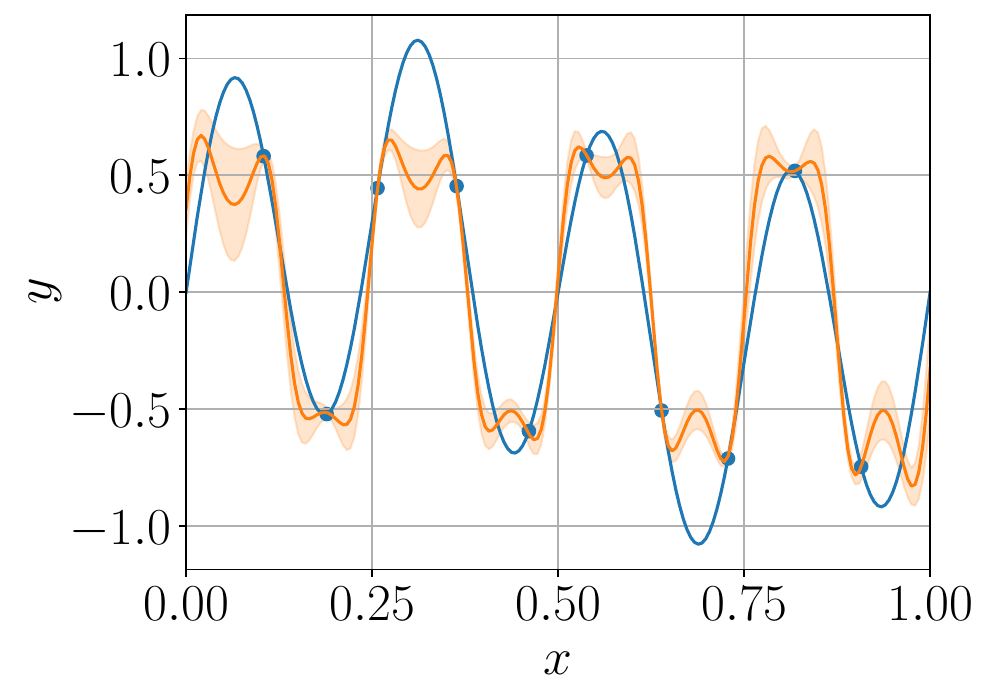}
        \caption{DGPDF}
        \label{fig:linear-dgpdf}
    \end{subfigure}
    \begin{subfigure}{.46\textwidth}
        \centering
        \includegraphics[width=\linewidth, height=0.18\textheight]{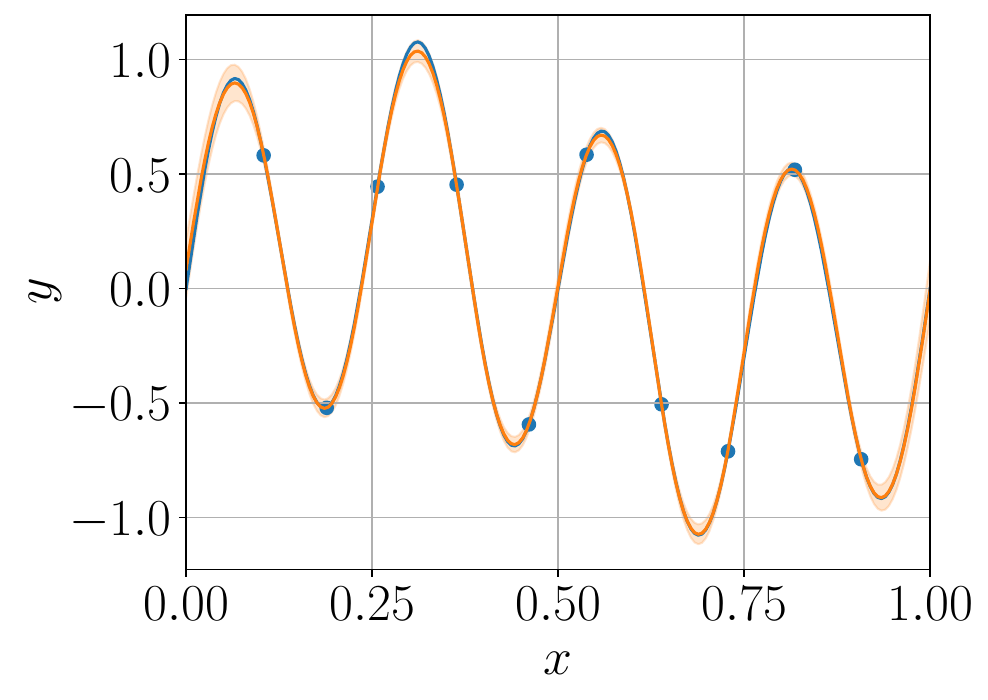}
        \caption{DGPDFC}
        \label{fig:linear-dgpdfc}
    \end{subfigure}
    \caption{Comparison of different multi-fidelity Gaussian Process surrogate modeling methods on a linear transformation problem stated in Table \ref{tab:academic-problems}. The plots show the high-fidelity function in the blue curve, the high-fidelity training data in blue dots, the predicted mean in the orange curve, and 95\% confidence interval using the orange-shaded region.}
    \label{fig:academic-linear}
\end{figure}

\begin{figure}[!]
    \begin{center}
        \includegraphics[width=.8\textwidth]{figures/legends/legend_predictions.pdf}
    \end{center}
    \begin{subfigure}{.46\textwidth}
        \centering
        \includegraphics[width=\linewidth, height=0.18\textheight]{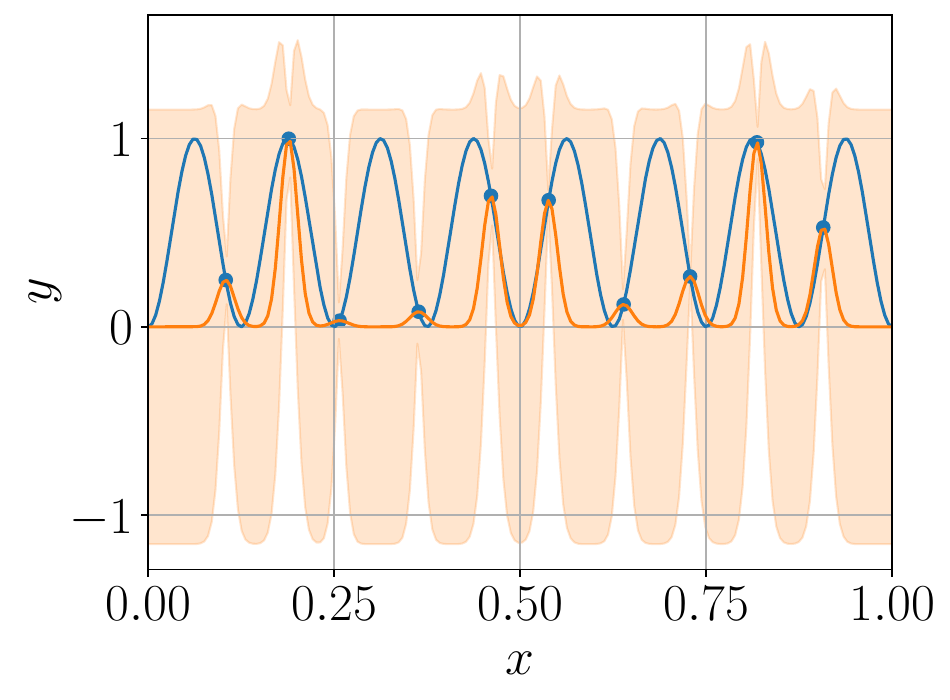}
        \caption{GP}
        \label{fig:nonlinear-gp}
    \end{subfigure} 
    \begin{subfigure}{.46\textwidth}
        \centering
        \includegraphics[width=\linewidth, height=0.18\textheight]{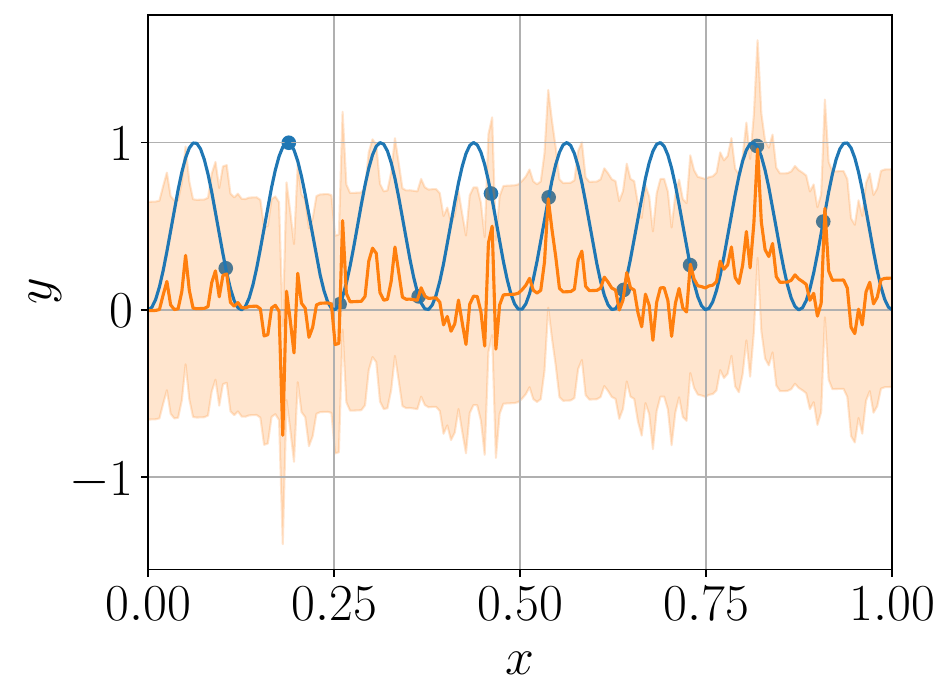}
        \caption{AR1}
        \label{fig:nonlinear-ar1}
    \end{subfigure} 
    \newline
    \begin{subfigure}{.46\textwidth}
        \centering
        \includegraphics[width=\linewidth, height=0.18\textheight]{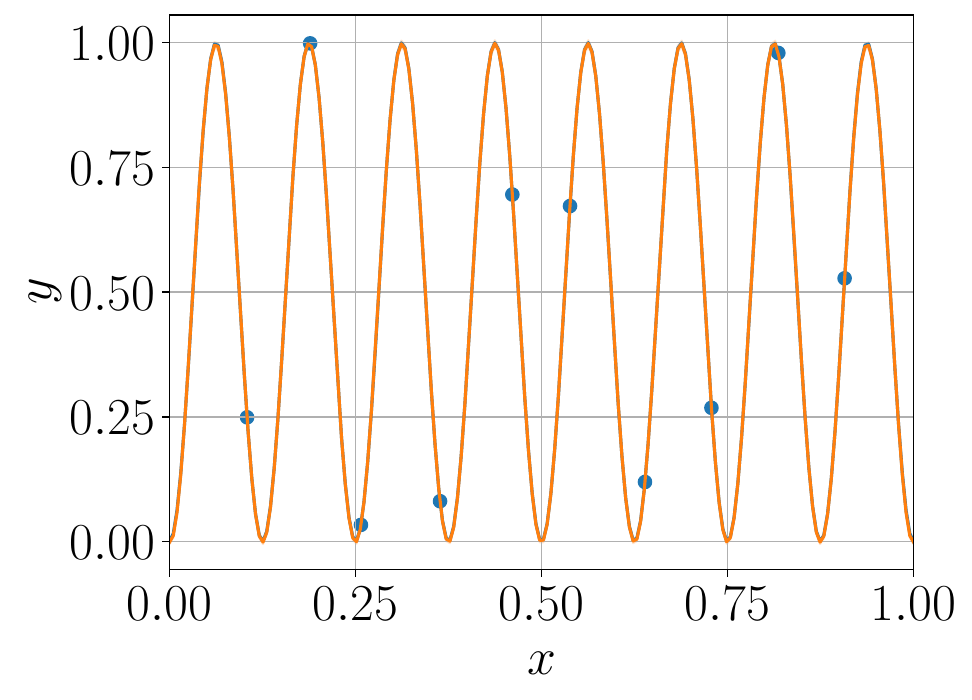}
        \caption{NARGP}
        \label{fig:nonlinear-nargp}
    \end{subfigure}
    \begin{subfigure}{.46\textwidth}
        \centering
        \includegraphics[width=\linewidth, height=0.18\textheight]{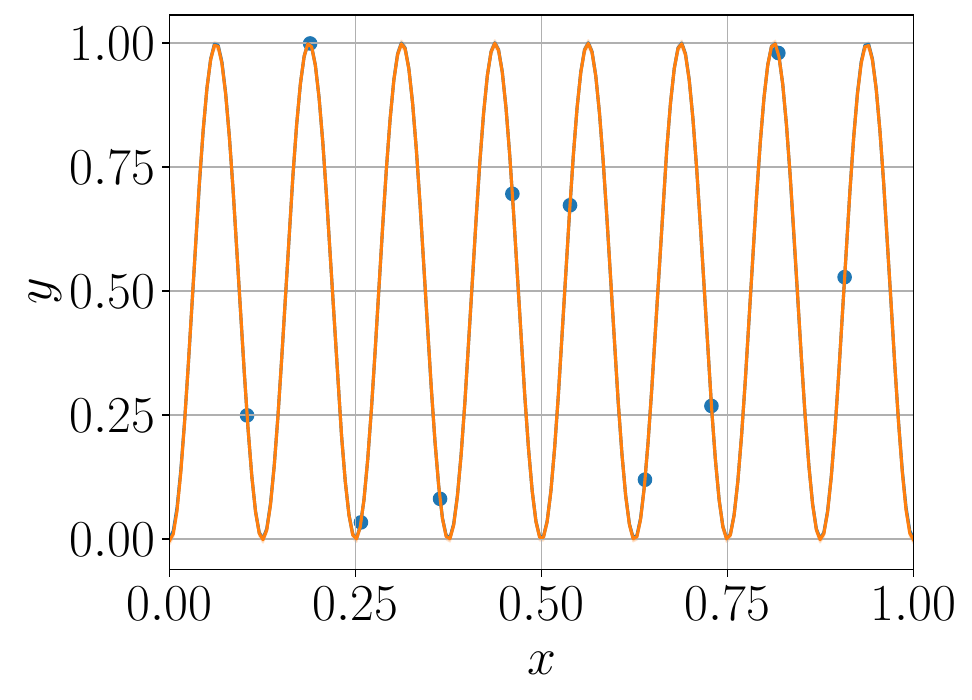}
        \caption{GPDF}
        \label{fig:nonlinear-gpdf}
    \end{subfigure}
    \newline 
    \begin{subfigure}{.46\textwidth}
        \centering
        \includegraphics[width=\linewidth, height=0.18\textheight]{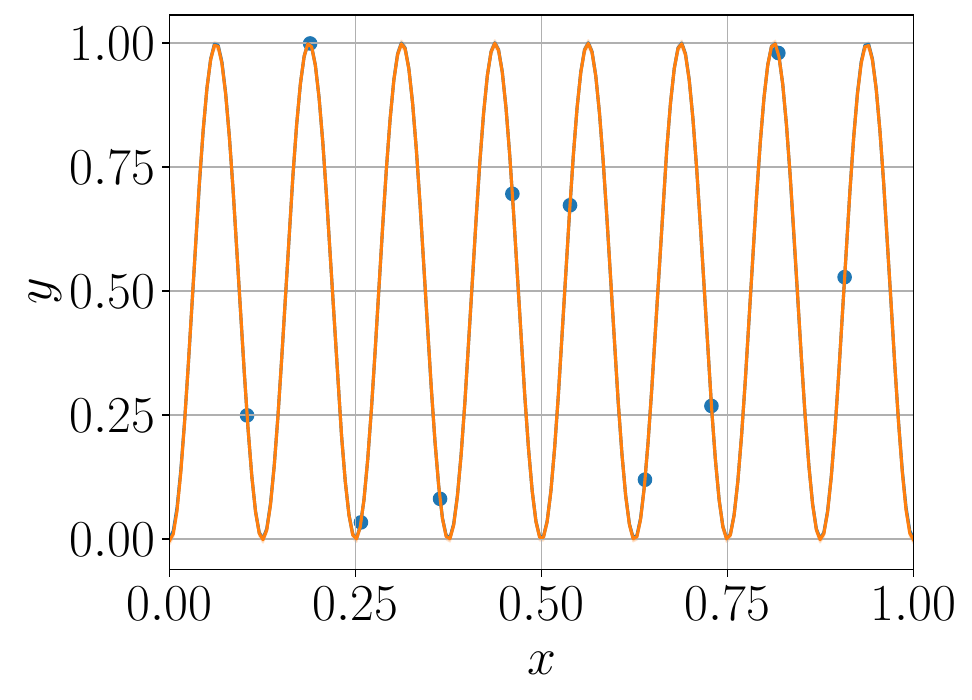}
        \caption{GPDFC}
        \label{fig:nonlinear-gpdfc}
    \end{subfigure}
    \begin{subfigure}{.46\textwidth}
        \centering
        \includegraphics[width=\linewidth, height=0.18\textheight]{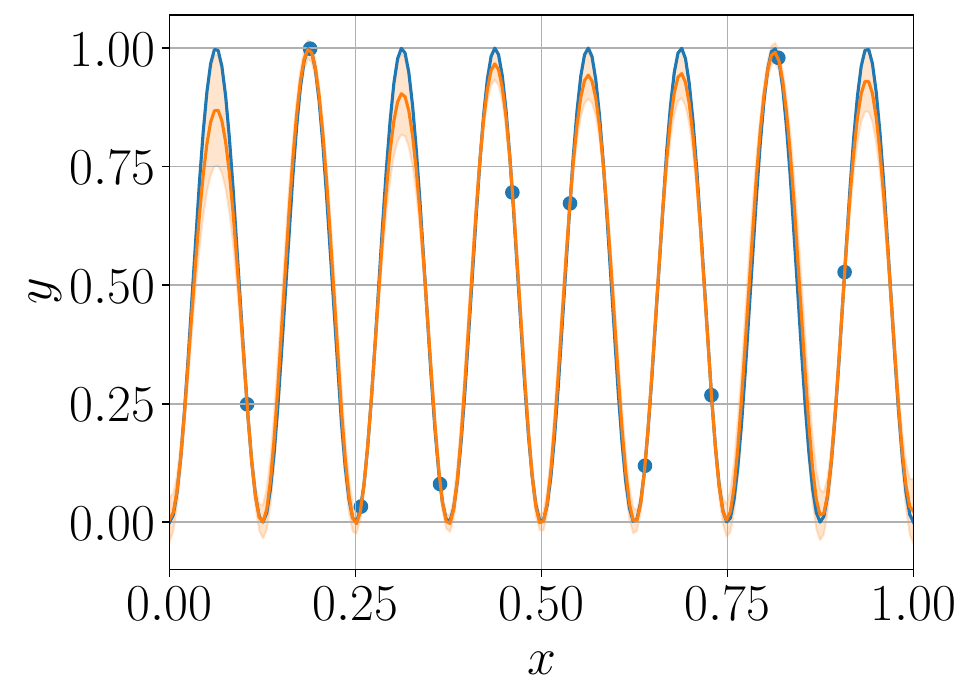}
        \caption{NARDGP}
        \label{fig:nonlinear-nardgp}
    \end{subfigure}
    \newline 
    \begin{subfigure}{.46\textwidth}
        \centering
        \includegraphics[width=\linewidth, height=0.18\textheight]{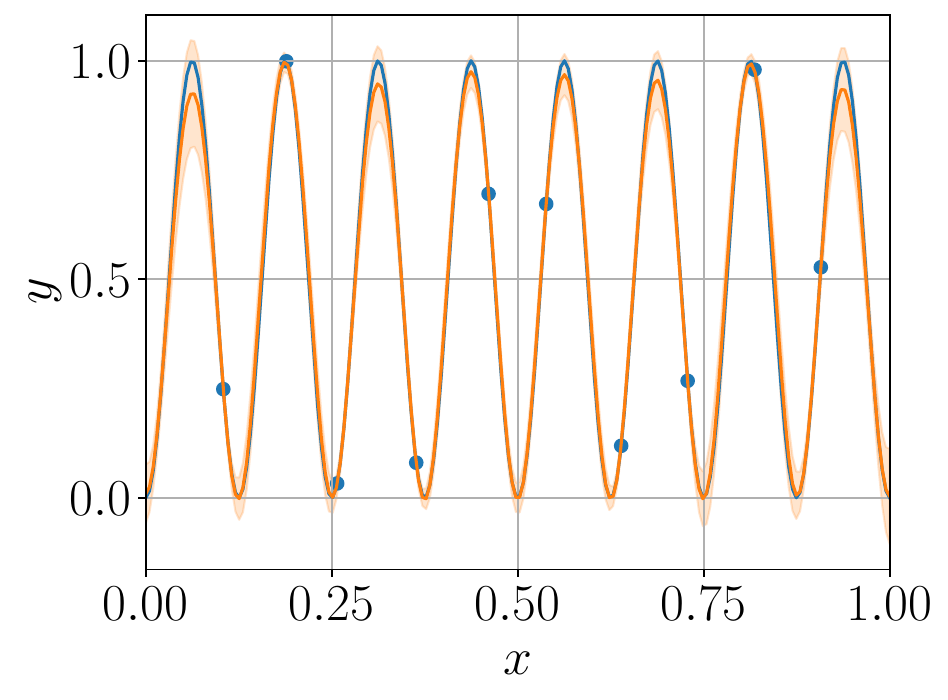}
        \caption{DGPDF}
        \label{fig:nonlinear-dgpdf}
    \end{subfigure}
    \begin{subfigure}{.46\textwidth}
        \centering
        \includegraphics[width=\linewidth, height=0.18\textheight]{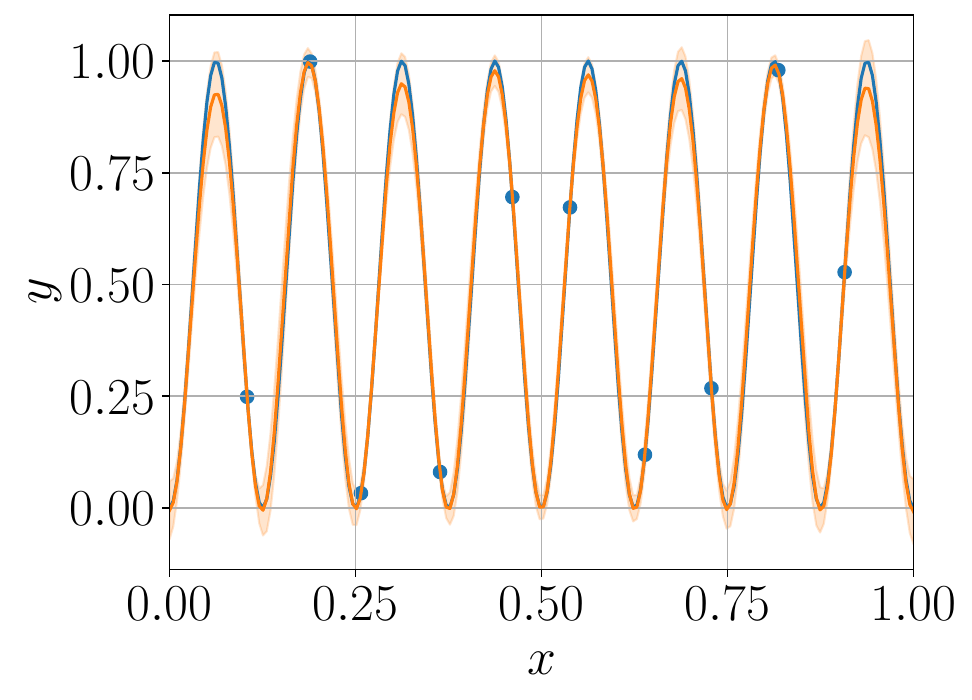}
        \caption{DGPDFC}
        \label{fig:nonlinear-dgpdfc}
    \end{subfigure}
    \caption{Comparison of different multi-fidelity Gaussian Process surrogate modeling methods on a non-linear transformation problem stated in Table \ref{tab:academic-problems}. The plots show the high-fidelity function in the blue curve, the high-fidelity training data in blue dots, the predicted mean in the orange curve, and 95\% confidence interval using the orange-shaded region.}
    \label{fig:academic-nonlinear}
\end{figure}

\begin{figure}[!]
    \begin{center}
        \includegraphics[width=.8\textwidth]{figures/legends/legend_predictions.pdf}
    \end{center}
    \begin{subfigure}{.46\textwidth}
        \centering
        \includegraphics[width=\linewidth, height=0.18\textheight]{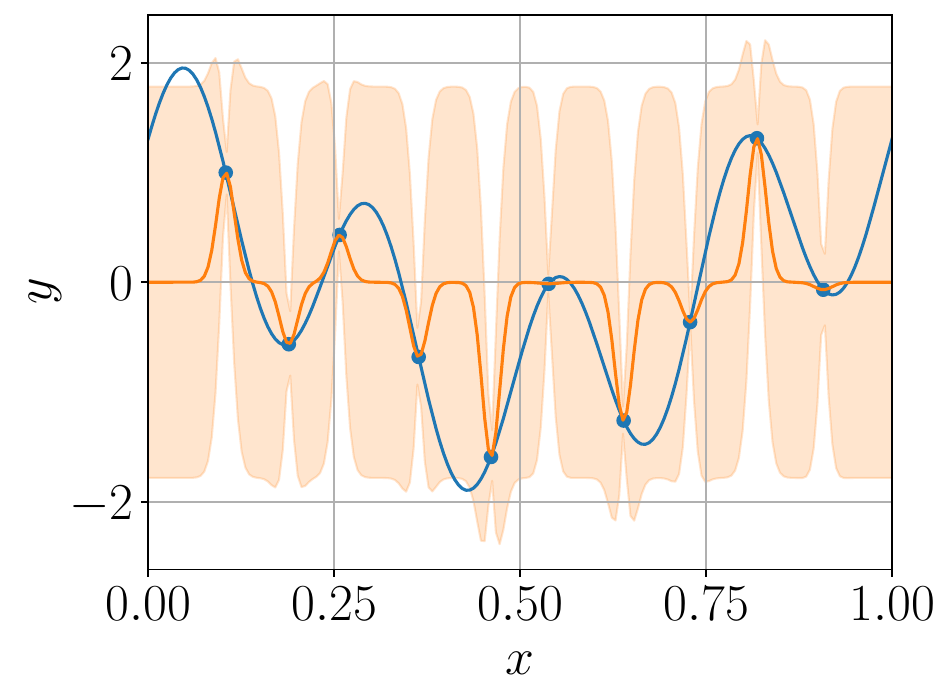}
        \caption{GP}
        \label{fig:phase_shift-gp}
    \end{subfigure}
    \begin{subfigure}{.46\textwidth}
        \centering
        \includegraphics[width=\linewidth, height=0.18\textheight]{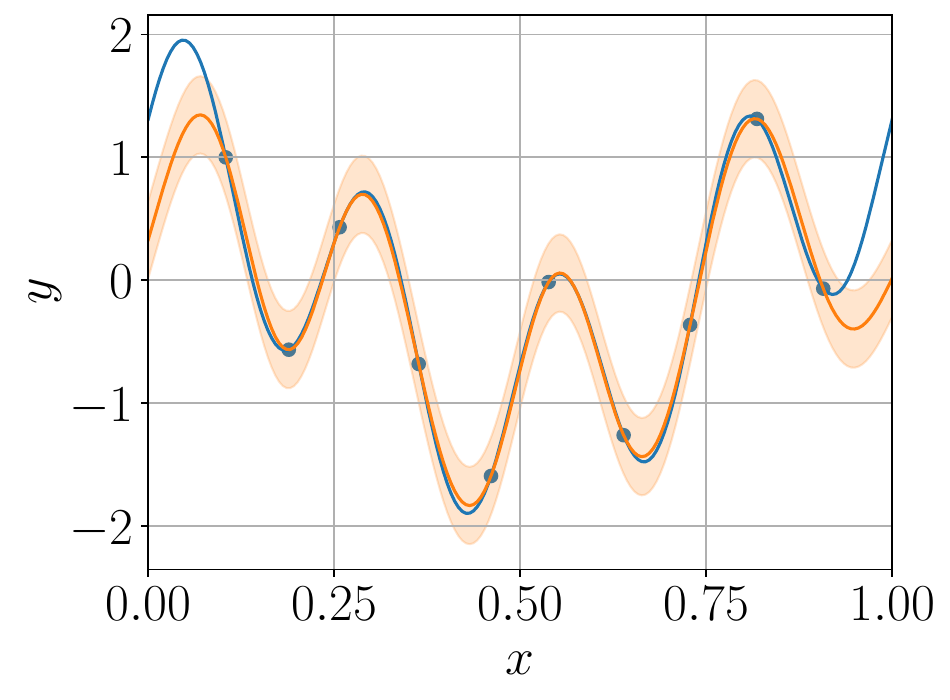}
        \caption{AR1}
        \label{fig:phase_shift-ar1}
    \end{subfigure} 
    \newline 
    \begin{subfigure}{.46\textwidth}
        \centering
        \includegraphics[width=\linewidth, height=0.18\textheight]{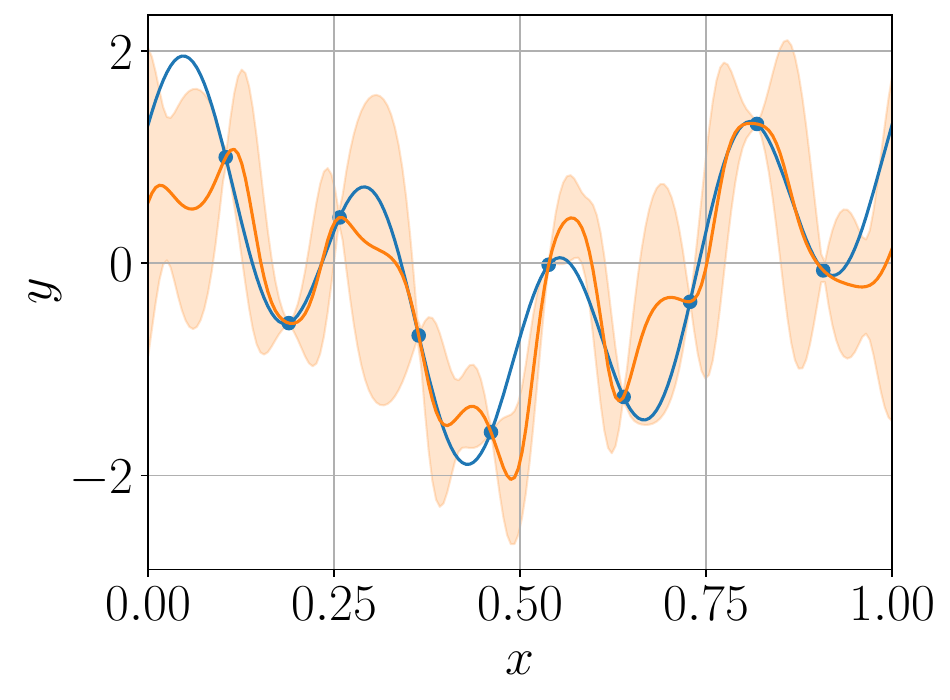}
        \caption{NARGP}
        \label{fig:phase-shift-nargp}
    \end{subfigure}
    \begin{subfigure}{.46\textwidth}
        \centering
        \includegraphics[width=\linewidth, height=0.18\textheight]{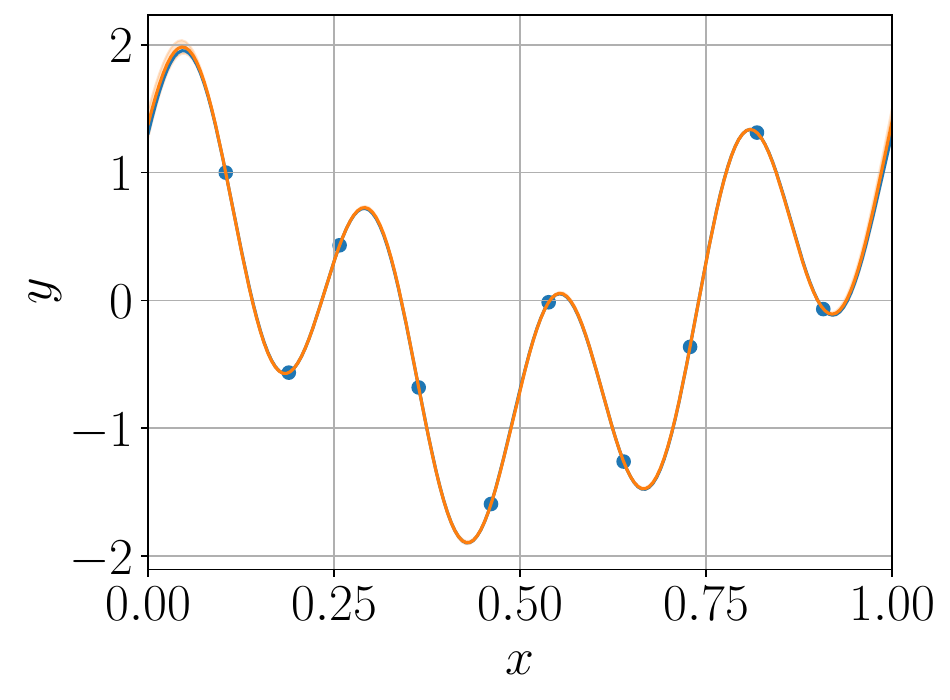}
        \caption{GPDF}
        \label{fig:phase-shift-gpdf}
    \end{subfigure}
    \newline 
    \begin{subfigure}{.46\textwidth}
        \centering
        \includegraphics[width=\linewidth, height=0.18\textheight]{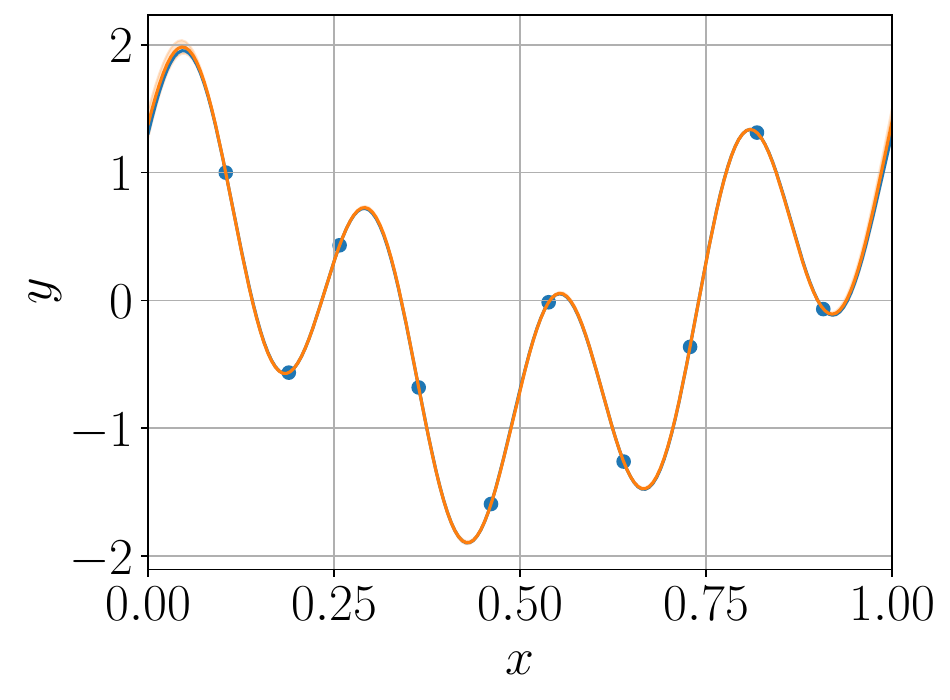}
        \caption{GPDFC}
        \label{fig:phase-shift-gpdfc}
    \end{subfigure}
    \begin{subfigure}{.46\textwidth}
        \centering
        \includegraphics[width=\linewidth, height=0.18\textheight]{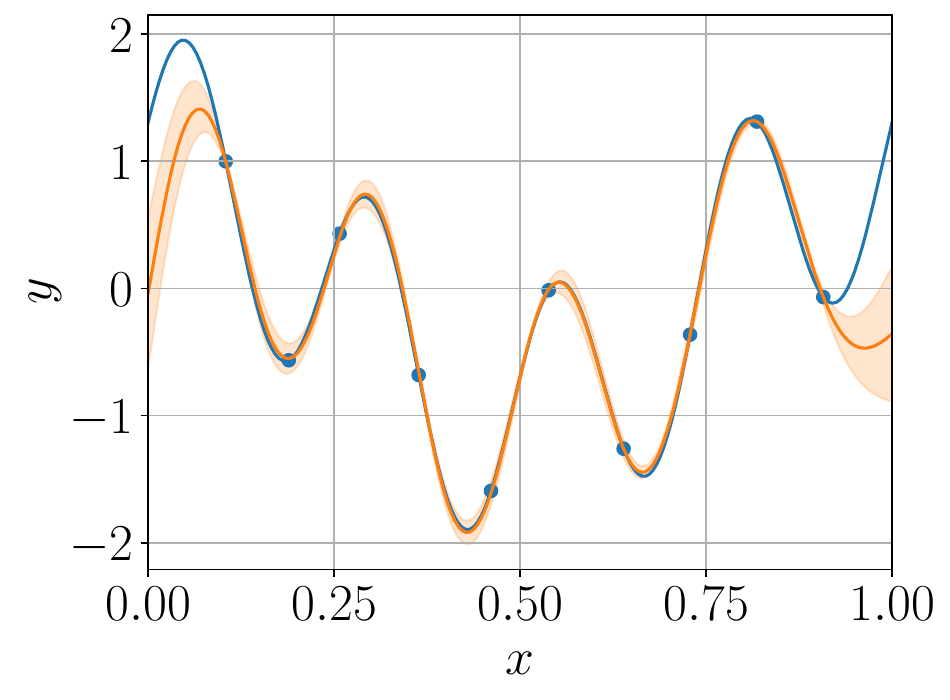}
        \caption{NARDGP}
        \label{fig:phase-shift-nardgp}
    \end{subfigure}
    \newline 
    \begin{subfigure}{.46\textwidth}
        \centering
        \includegraphics[width=\linewidth, height=0.18\textheight]{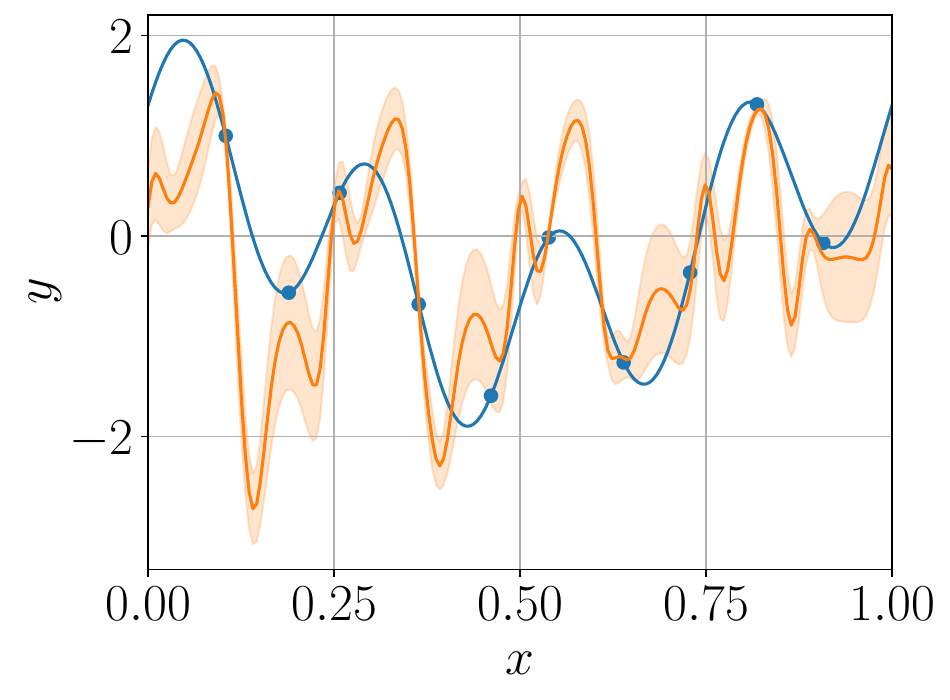}
        \caption{DGPDF}
        \label{fig:phase-shift-dgpdf}
    \end{subfigure}
    \begin{subfigure}{.46\textwidth}
        \centering
        \includegraphics[width=\linewidth, height=0.18\textheight]{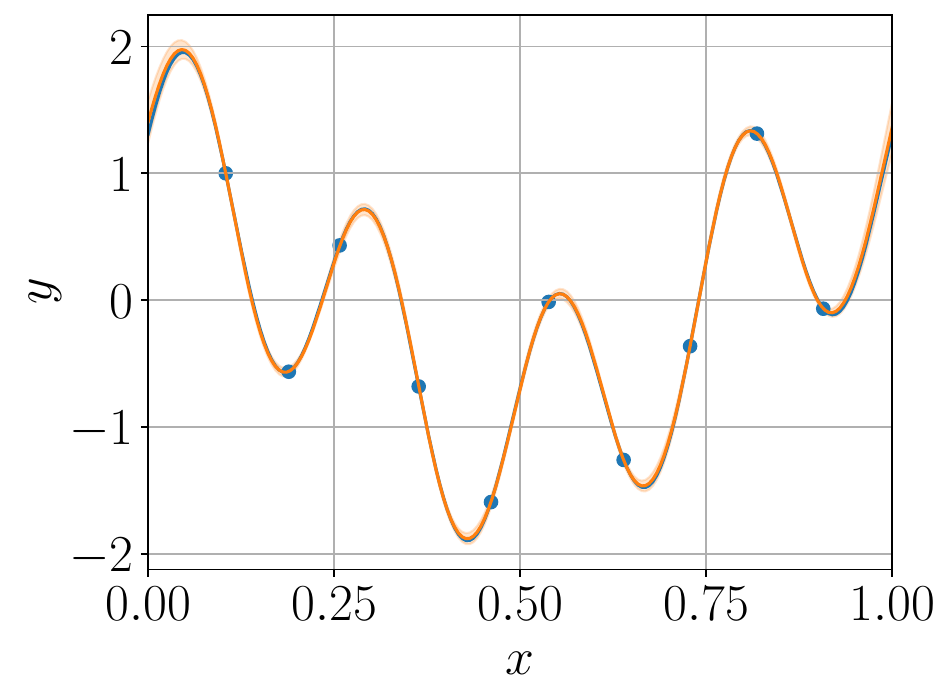}
        \caption{DGPDFC}
        \label{fig:phase-shift-dgpdfc}
    \end{subfigure}
    \caption{Comparison of different Multi-fidelity Gaussian Process surrogate modeling methods on a phase-shifted oscillation stated in Table \ref{tab:academic-problems}. The plots show the high-fidelity function in the blue curve, the high-fidelity training data in blue dots, the predicted mean in the orange curve, and 95\% confidence interval using the orange-shaded region.}
    \label{fig:academic-phase-shift}
\end{figure}